\documentclass[journal]{IEEEtran}
\usepackage{amssymb} 
\usepackage{amsmath}
\usepackage{upgreek}
\usepackage{bm}
\usepackage[switch]{lineno}
\usepackage{textcomp,mathcomp}
\usepackage{graphicx}  
\usepackage{float}  
\usepackage{subfigure}  
\usepackage{booktabs}
\usepackage{multirow}
\usepackage{colortbl} 
\usepackage{xcolor}
\usepackage{threeparttable}

\usepackage[linesnumbered,ruled,lined]{algorithm2e}
\definecolor{tablecolor0}{RGB}{224,224,224}
\definecolor{tablecolor1}{RGB}{235, 233, 242}
\definecolor{tablecolor2}{RGB}{213,209,228}
\definecolor{tablecolor3}{RGB}{241,221,230}
\definecolor{tablecolor4}{RGB}{239, 239, 239}
\definecolor{tablecolor5}{RGB}{236,228,213}
\definecolor{tablecolor6}{RGB}{136, 149, 177}

\definecolor{tablecolor7}{RGB}{214,180,179}
\definecolor{tablecolor8}{RGB}{245,236,237}
\definecolor{tablecolor9}{RGB}{193,211,199}
\definecolor{tablecolor10}{RGB}{243,244,239}
\definecolor{tablecolor11}{RGB}{249,220,178}
\definecolor{tablecolor12}{RGB}{251,236,195}
\definecolor{tablecolor13}{RGB}{245,239,255}
\definecolor{tablecolor14}{RGB}{237,241,245}
\begin{document}

\title{Nearshore Underwater Target Detection Meets UAV-borne Hyperspectral Remote Sensing: \\A Novel Hybrid-level Contrastive Learning Framework and Benchmark Dataset}

\author{\IEEEauthorblockN{Jiahao~Qi, Chuanhong Zhou, Xingyue~Liu, Chen~Chen, Dehui Zhu, Kangcheng Bin and Ping~Zhong~\IEEEmembership{Senior Member,~IEEE}}
\thanks{This work was supported in part by the Foundation Fund of Science and Technology on Near-Surface Detection Laboratory under Grant 6142414220808, and in part by the National Natural Science Foundation of China 62201586, and in part by China National Postdoctoral Program for Innovative Talents under Grant BX20240492. \emph{(Corresponding author: Ping Zhong.)}}
\thanks{Jiahao Qi, Xingyue Liu, Chen Chen and Ping Zhong are with the National Key Laboratory of Science and Technology on Automatic Target Recognition, National University of Defense Technology, Changsha 410073, China (e-mail: qijiahao1996@nudt.edu.cn, xingyueliu0801@nudt.edu.cn, chenchen21c@nudt.edu.cn, zhongping@nudt.edu.cn).}
}

\markboth{Submitted to IEEE Transactions on Geoscience and Remote Sensing}%
{Shell \MakeLowercase{\textit{et al.}}: A Sample Article Using IEEEtran.cls for IEEE Journals}


\maketitle

\begin{abstract}
    UAV-borne hyperspectral remote sensing has emerged as a promising approach for underwater target detection (UTD). 
    However, its effectiveness is hindered by spectral distortions in nearshore environments, which compromise the accuracy of traditional hyperspectral UTD (HUTD) methods that rely on bathymetric model. 
    These distortions lead to significant uncertainty in target and background spectra, challenging the detection process.
    To address this, we propose the Hyperspectral Underwater Contrastive Learning Network (HUCLNet), a novel framework that integrates contrastive learning with a self-paced learning paradigm for robust HUTD in nearshore regions. HUCLNet extracts discriminative features from distorted hyperspectral data through contrastive learning, while the self-paced learning strategy selectively prioritizes the most informative samples. Additionally, a reliability-guided clustering strategy enhances the robustness of learned representations.
    To evaluate the method effectiveness, we conduct a novel nearshore HUTD benchmark dataset, ATR2-HUTD, covering three diverse scenarios with varying water types and turbidity, and target types. Extensive experiments demonstrate that HUCLNet significantly outperforms state-of-the-art methods. 
    The dataset and code will be publicly available at: \emph{https://github.com/qjh1996/HUTD}.    
\end{abstract}

\section{Introduction}\label{sec:1} 
\IEEEPARstart{U}{nderwater} target detection (UTD)~\cite{Liu2024,Zhang2023,Li2023} aims to locate and identify underwater objects, providing essential data for ecosystem conservation and sustainable resource management to mitigate environmental threats.  
Despite its significance, effective UTD in nearshore regions remains challenging due to the dynamic and complex underwater environment, necessitating rapid, large-scale data acquisition.  
Remote sensing~\cite{8697135, 9174822} offers a promising solution by enabling extensive spatial data collection with high temporal resolution.  
\par
\begin{figure}[!t]                 
    \centering                    
    \includegraphics[width=1\columnwidth]{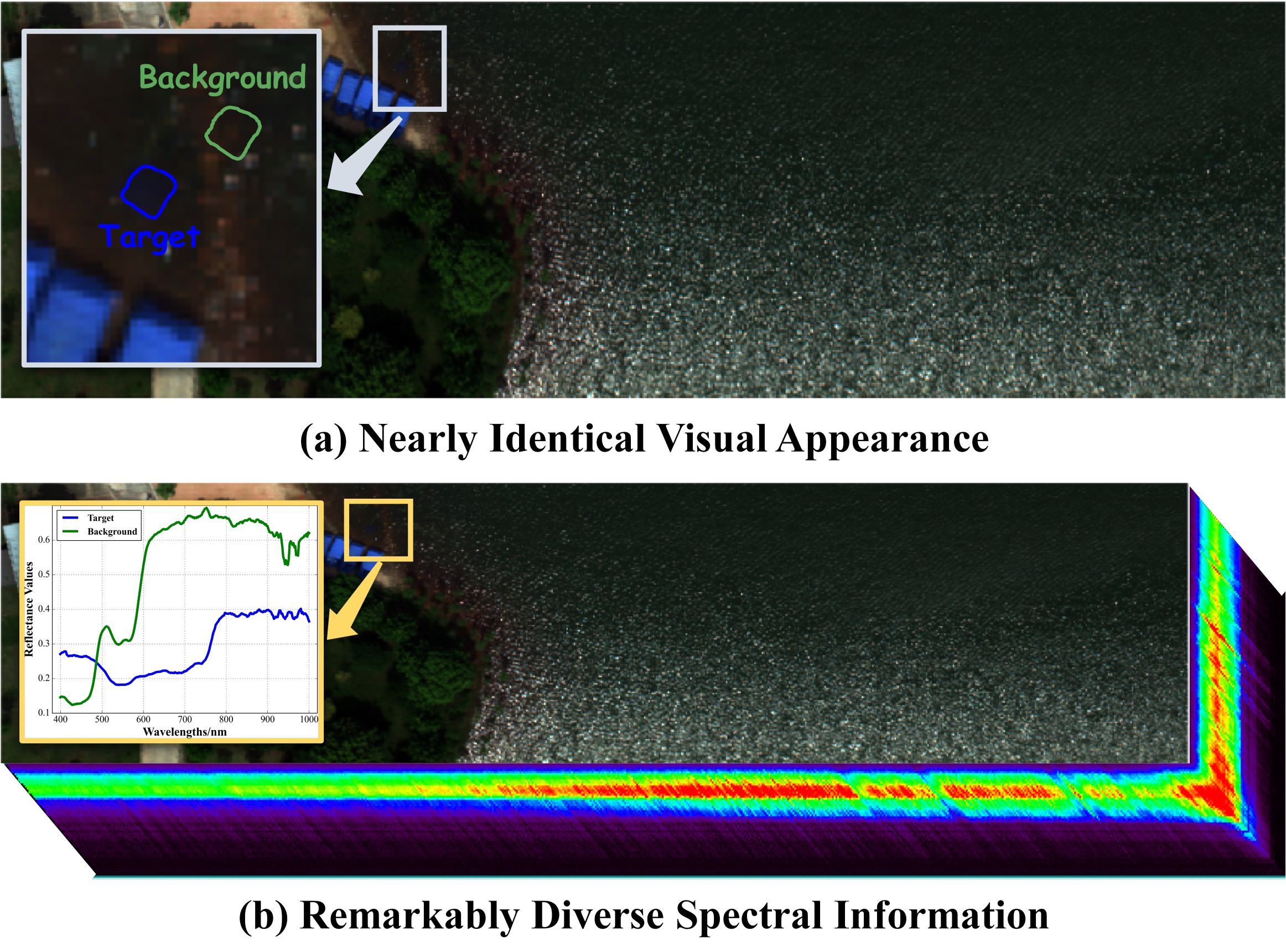}                     
    \caption{Limitations of RGB imagery and advantages of hyperspectral imagery in underwater depth estimation. (a) In RGB images, the target and background have nearly identical spatial appearances; (b) In hyperspectral images, the target and background exhibit distinct spectral signatures.}                  
    \label{fig:A0}   
\end{figure} 
\begin{figure*}[!t]                 
    \centering                    
    \includegraphics[width=2\columnwidth]{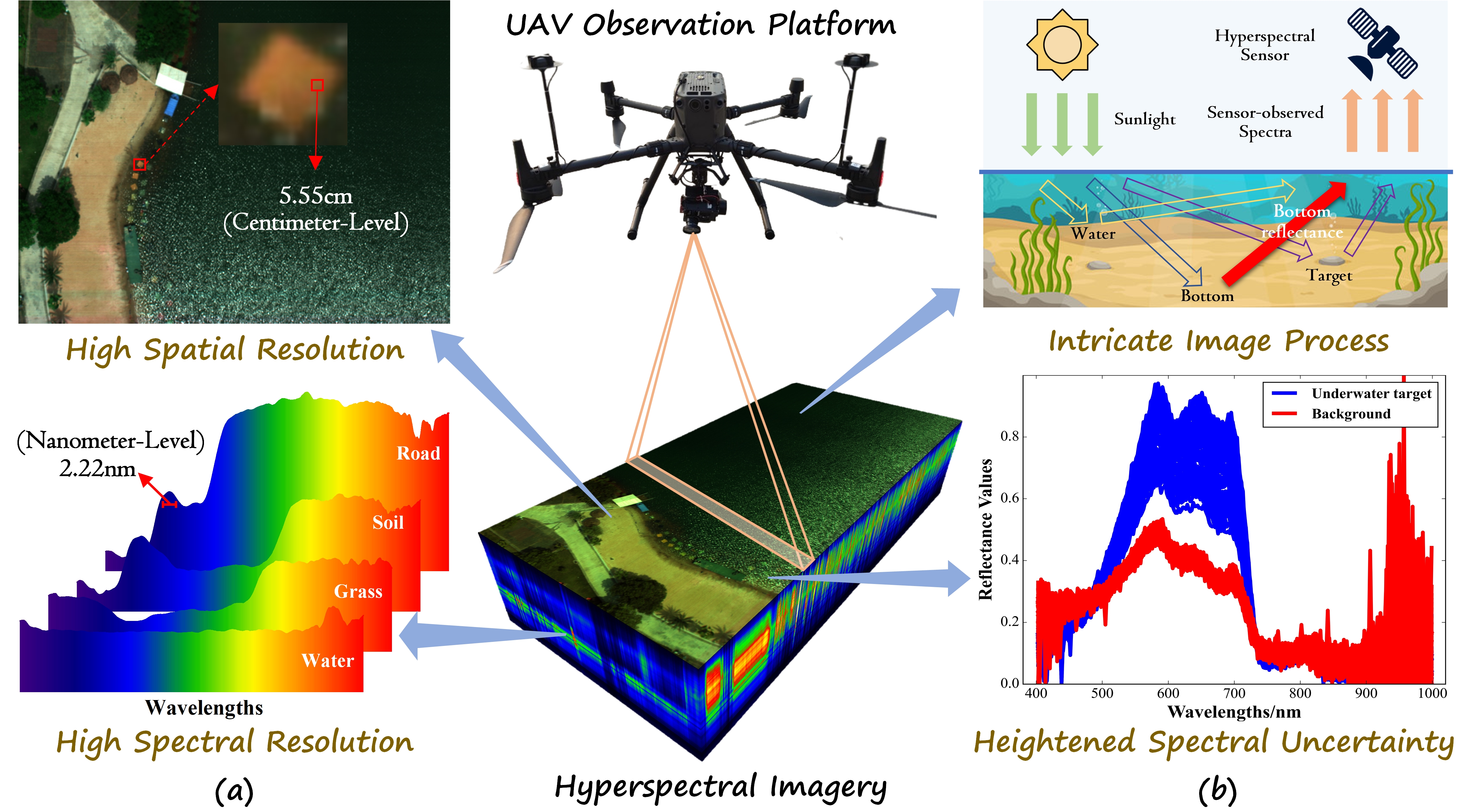}                     
    \caption{The illustration of opportunities and challenges in hyperspectral nearshore underwater target detection. (a) Opportunities; (b) Challenges.}                  
    \label{fig:A1}    
\end{figure*} 
RGB images are the most commonly used data type in remote sensing and play a crucial role in environmental monitoring~\cite{Ma2019}.  
They capture spatial features such as texture, shape, and color, which are essential for general environmental analysis.  
However, RGB imagery has significant limitations in nearshore UTD.  
Studies~\cite{10704737,10336777} show that radiation at 0.45$\upmu$m (blue) and 0.65$\upmu$m (red) is strongly absorbed by chlorophyll, reducing reflectance in these bands and limiting the capture of underwater scene details.  
Consequently, RGB-based spatial features often lack discriminability in nearshore environments.  
As illustrated in Fig.~\ref{fig:A0} (a), this limitation is further exacerbated by the restricted spatial resolution of remote sensing data, which reduces the distinctiveness of underwater targets against the background.  
Similar constraints exist in other spatial imaging modalities, such as infrared imagery, suggesting that spatial features alone are insufficient for nearshore UTD.  
\par  
In contrast, hyperspectral imagery (HSI) captures hundreds to thousands of narrow spectral bands, providing rich spectral information across the visible, near-infrared, and shortwave infrared regions.  
This enables precise target identification based on unique spectral signatures, even in complex optical conditions~\cite{Manolakis2014}.  
Unlike RGB imagery, which relies on spatial features, HSI offers fine-grained spectral features that enhance target-background differentiation.  
Its extensive spectral coverage mitigates water absorption effects, facilitating accurate modeling of underwater scenes.  
As shown in Fig.~\ref{fig:A0} (b), underwater targets and background regions exhibit distinct spectral characteristics despite their similar spatial appearances, making HSI well-suited for nearshore UTD.  
\par  
Recent advances in hyperspectral remote sensing have been driven by satellite, airborne, and UAV-based platforms.  
Among these, UAV-based HSI systems offer significant advantages for nearshore UTD, as illustrated in Fig.~\ref{fig:A1} (a).  
They provide high spatial resolution imagery, often at the centimeter scale, minimizing subpixel interference~\cite{Jiao2022}.  
Additionally, UAV-acquired hyperspectral data are less affected by atmospheric attenuation and environmental noise, ensuring higher image quality than other platforms~\cite{Phang2023, Gu2023}.  
With their flexibility, cost-effectiveness, and real-time data acquisition capabilities, UAVs are particularly suited for monitoring dynamic nearshore environments~\cite{Zhong2020}.  
These attributes position UAV-based HSI as a promising solution for addressing UTD challenges, forming the focus of this study.  
\subsection{UAV-borne Hyperspectral Underwater Target Detection}
\begin{figure}[!t]                  
    \centering                    
    \includegraphics[width=1\columnwidth]{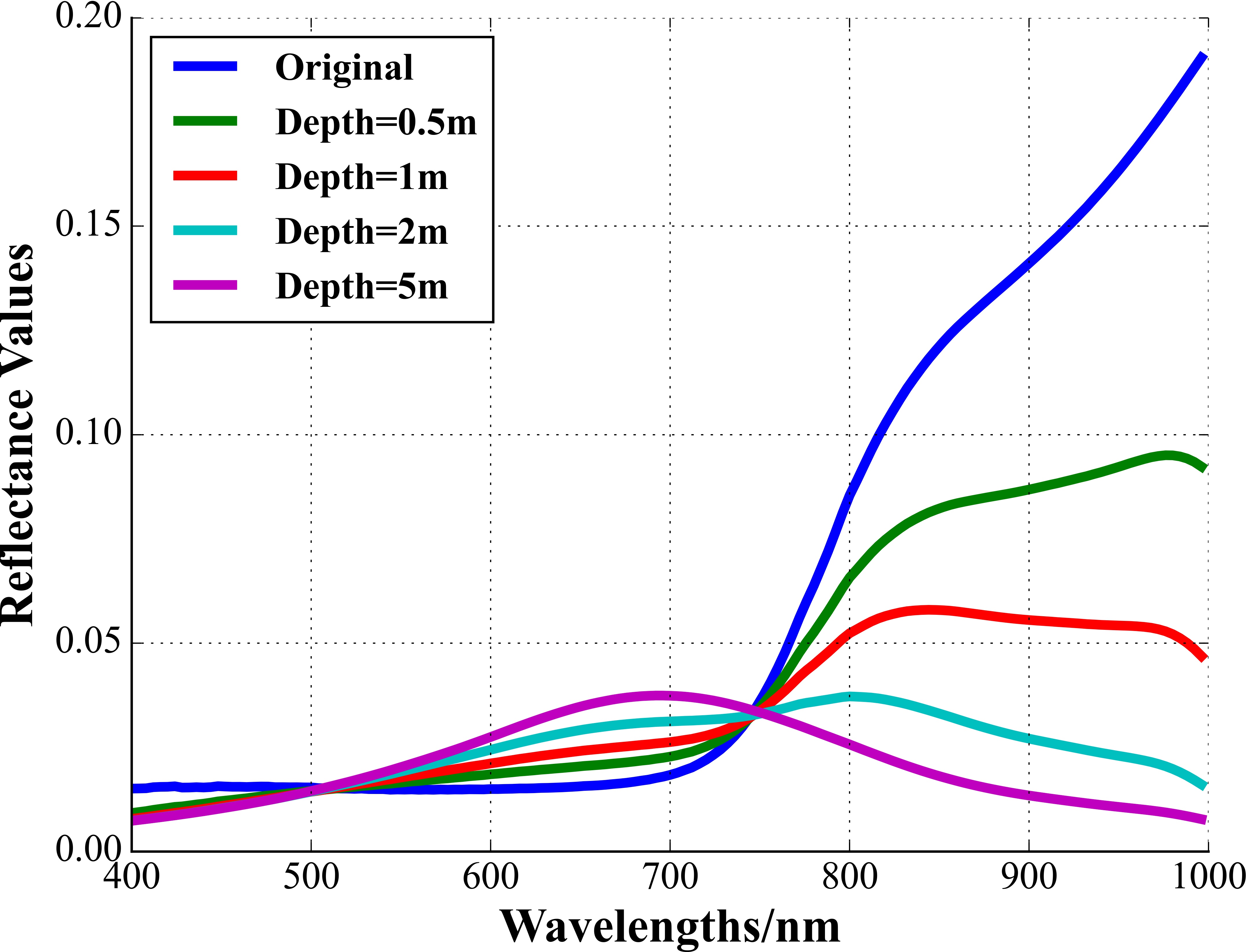}                     
    \caption{Illustration of spectral distortions induced by underwater conditions, with depth as an example. 
    The spectral signature of underwater target diverge from their reference spectra, and this deviation varies with depth.
    The similar situations can be observed for other underwater conditions~\cite{Gillis2020}.}           
    \label{fig:A1-2}    
\end{figure}

Despite the advantages of hyperspectral target detection (HTD), its application in underwater environments is hindered by spectral distortions induced by the water column~\cite{Gillis2020}.  
Light absorption and scattering alter the spectral signatures of underwater targets, causing deviations from their reference spectra\footnote{
    Reference spectra denote the known spectral signatures of underwater targets, obtained either on land or in controlled settings.
}~\cite{Gillis2020}.  
Existing HTD methods assume that target spectra match their reference spectra~\cite{Manolakis2014}, an assumption frequently violated in underwater conditions.  
As shown in Fig.~\ref{fig:A1-2}, spectral distortions vary with depth, turbidity, and water composition, further degrading the accuracy of conventional HTD approaches.  
To address these challenges, prior studies~\cite{Liu2024,Gillis2020,Jay2012,LiZheyong2023,Qi2021} have explored two primary detection strategies.  
\par
The first strategy predicts underwater spectral signatures from known reference spectra using a bathymetric model~\cite{Lee1998}, followed by target detection.  
Jay \emph{et al.}~\cite{Jay2012} propose a method that corrects water column distortions using a bathymetric model and detects targets with a GLRT-based adaptive filter, without requiring prior water parameter knowledge.  
Similarly, Gillis~\cite{Gillis2020} develops a framework that predicts submerged target spectra through radiative transfer modeling and nonlinear dimensionality reduction.  
More recently, Li \emph{et al.}~\cite{LiZheyong2023} introduce a transfer-based hyperspectral underwater target detection (HUTD) framework that synthesizes spectral data at various depths and employs domain adaptation for improved target detection.  

\par  
The second strategy restores reference spectra from observed underwater spectra using hyperspectral unmixing techniques.  
This approach is also based on the bathymetric model, which expresses the observed underwater spectrum as a linear combination of the reference spectrum and the water body spectrum.  
Qi \emph{et al.}~\cite{Qi2021} develop the first unmixing-based HUTD network, reconstructing reference spectra through hyperspectral unmixing.  
Liu \emph{et al.}~\cite{Liu2024} extend this approach with a nonlinear representation, adapting unmixing-based HUTD for nearshore scenarios.  

\par  
Despite promising results, several challenges remain in adapting these methods to nearshore environments:  
\begin{itemize}  
    \item \textbf{Dependency on the bathymetric model.}  
    Existing HUTD methods rely on the bathymetric model to describe underwater imaging mechanisms.  
    However, its assumptions, including linear mixing and uniform water column properties~\cite{Lee1998}, often do not reflect real-world conditions.  
    These limitations, illustrated in Fig.~\ref{fig:A1} (b, Top), undermine detection accuracy and generalizability.  
    Furthermore, effective use of the bathymetric model requires prior knowledge of target depth, suspended particle concentration, and water optical properties~\cite{Gillis2020}, which are challenging to obtain in dynamic nearshore environments.  

    \item \textbf{Limited target characterization.}  
    As shown in Fig.~\ref{fig:A1} (b, Bottom), spectral signatures of the same material vary significantly in nearshore environments due to factors such as water quality, turbidity, and light attenuation.  
    This spectral variability increases uncertainty, degrading the performance of existing HUTD methods.  
    Prediction-based approaches struggle with fluctuating environmental parameters, while restoration-based methods face additional distortions that complicate spectral unmixing.  
    These challenges stem from an overemphasis on spectral restoration or prediction rather than accurate target characterization, which is crucial for nearshore UTD.  
    In dynamic nearshore conditions, where spectral differences between targets and backgrounds are minimal, precise target characterization is essential for improving detection performance.  
\end{itemize}  

\subsection{UAV-borne Hyperspectral Underwater Target Detection Datasets}
Advancements in hyperspectral remote sensing depend on high-quality data.  
The performance of HUTD algorithms is strongly influenced by the diversity and comprehensiveness of training and evaluation datasets.  
UAV-based hyperspectral datasets are essential for assessing detection algorithms under realistic underwater conditions.  
Several HUTD datasets have been introduced, each contributing unique insights.  
\par  
Zhang \emph{et al.}~\cite{Zhang2023} collected two HUTD datasets using a Headwall Nano-Hyperspec sensor mounted on a DJI Matrice 600 Pro UAV at 40 m altitude over Qingdao and Liaocheng, China.  
The HNU-UTD dataset includes Tetrapods, Cement, and Plants as underwater targets.  
Li \emph{et al.}~\cite{LiZheyong2023} introduced the NPU-Pool dataset, acquired with a Gaia Field-V10 imager spanning 400--1000 nm and a spatial resolution of $100 \times 100$, with targets at depths of 0 to 3.1 m under controlled indoor lighting.  
Meanwhile, Li \emph{et al.}~\cite{Li2023} proposed the NPU-Sea dataset, captured in Sanya, Hainan Province, under real seawater conditions, where sea surface waves, water quality, and target movement influenced the data.  
This dataset includes iron plate targets at depths of 0.8 m and 3.0 m.  
More recently, Liu \emph{et al.}~\cite{Liu2024} introduced the ATR2-Lake dataset, collected at Qianlu Lake Reservoir, China, using a Headwall Nano-Hyperspec sensor mounted on a DJI Matrice 300 RTK UAV.  
The dataset includes black metal plates at depths of 1 to 3 m.  
Tab.~\ref{dataset} summarizes the main characteristics of these datasets.
\begin{table*}[!t] 
    \centering
    \renewcommand{\arraystretch}{2}
    \setlength{\tabcolsep}{1.9mm}
    \caption{Key characteristics of existing hyperspectral underwater target detection datasets.} \label{dataset}
    \begin{threeparttable}
    \scalebox{0.9}
    {   
        \begin{tabular}{ccccccccc}
            \toprule
            \multirow{2}{*}{\textbf{Dataset}} & \multicolumn{2}{c}{\cellcolor{tablecolor1}\textbf{Sensor-Related}} & \multicolumn{3}{c}{\cellcolor{tablecolor2}\textbf{Dataset-Related}}                                               & \multicolumn{2}{c}{\cellcolor{tablecolor3}\textbf{Target-Related}}       \\ \cmidrule(lr){2-3} \cmidrule(lr){4-6} \cmidrule(lr){7-8}
                                     & Wavelength  & Spectral Resolution  & Image Size                   & Scenario Type    & Accessible & Target Type               & Target Depth \\ \midrule
            \textbf{HNU-UTD}$^1$                  & 400-1000nm  & 2.2nm    & 560$\times$610, 250$\times$250   & Sea              & Yes        & Tetrapod, Cement, Plants & Unknown      \\
            \textbf{NPU-Pool}$^{2,3}$                 & 400-780 nm  & 3.5nm                & 100$\times$100                      & Anechoic pool    & No         & Iron, Stone, Rubber       & 0-3.1m       \\
            \textbf{NPU-Sea}$^3 $                  & 400-780 nm  & 3.5nm                & 350$\times$350                      & Sea              & No         & Iron                      & 0.8m, 3m     \\
            \textbf{ATR2-Lake}               & 400-1000nm  & 2.2nm               & 242$\times$341, 255$\times$261, 137$\times$178    & Lake             & Yes        & Metal                     & 1m-3m        \\
            \rowcolor{tablecolor5!50}\textbf{ATR2-HUTD}                & 400-1000nm  & 2.2nm               & 2304$\times$640, 3536$\times$640, 3171$\times$640 & Sea, Lake, River & Yes        & Metal, Wooden, Plastic    & 1m-3m       \\ \bottomrule
            \end{tabular}
    }
    \begin{tablenotes}
        \item[1] HNU-UTD dataset collected HSIs via both UAV and Satellite platforms, but we only list the UAV-related data. 
        \item[2] NPU-Pool dataset includes both outdoor and indoor sub-datasets, but the original paper only details the indoor sub-dataset, with no information on the \\ outdoor sub-dataset. As the dataset is not publicly accessible, only the indoor sub-dataset is included in the table.
        \item[3] NPU-Pool and NPU-Sea datasets were collected using tripods rather than UAVs, but are included here for comparison due to the limited availability of \\ HUTD datasets.
    \end{tablenotes}
    \end{threeparttable}
\end{table*}
\par  
Despite their contributions, existing datasets have critical limitations that hinder the development of robust HUTD algorithms.  
Based on Tab.~\ref{dataset}, the key limitations can be summarized as follows:  
\begin{itemize}  
    \item \textbf{Limited data scale.}  
    Most HUTD datasets are relatively small, typically comprising only hundreds of pixels.  
    This restricted scale fails to capture the complexity of underwater scenes, limiting background variability and hindering generalization to diverse nearshore environments.  
    It also constrains the thorough evaluation of detection algorithms.  
    \item \textbf{Insufficient scene diversity.}  
    Many datasets focus on specific water types, such as seas~\cite{Zhang2023}, lakes~\cite{Liu2024}, or controlled environments~\cite{LiZheyong2023}, failing to represent the full optical variability of real-world underwater conditions.  
    Differences in turbidity, salinity, and light attenuation remain underrepresented, leading to potential overfitting and reduced model adaptability in dynamic aquatic environments.  
\end{itemize}  
\par
\subsection{Contributions of This Study}  
In this paper, we introduces a novel contrastive learning framework, \textbf{H}yperspectral \textbf{U}nderwater \textbf{C}ontrastive \textbf{L}earning \textbf{N}etwork (HUCLNet), which integrates a self-paced learning (SPL) paradigm to address key challenges in HUTD.  
Unlike conventional prediction- or restoration-based approaches, HUCLNet learns a semantically rich latent space, where underwater target spectra are closely aligned with reference spectra while remaining distinct from background spectra in a data-driven manner.  
\par
HUCLNet comprises two core modules: the reliability-guided clustering (RGC) module and the hybrid-level contrastive learning (HLCL) module.  
The RGC module assigns hyperspectral pixels to prototypes via unsupervised clustering, incorporating a fixed prototype derived from the reference spectrum.  
A novel reliability criterion is introduced to assess cluster trustworthiness, refining pixel assignments into reliable clusters and unreliable instances.  
The HLCL module processes unreliable instances via instance-level contrastive learning to enhance discriminative representation and clustering accuracy, while reliable clusters undergo prototype-level contrastive learning to align target spectra with references while maintaining separation from background spectra.  
To further enhance contrastive learning, we propose a hyperspectral-specific data augmentation strategy based on unsupervised adversarial training.  
The entire framework follows the SPL paradigm, progressively incorporating unreliable instances into reliable clusters as the HLCL module improves target characterization, thereby strengthening representation learning and improving HUTD performance.  
Experimental results demonstrate that HUCLNet significantly outperforms state-of-the-art (SOTA) HUTD methods, effectively addressing key methodological gaps in the field.  
\par
Beyond the proposed framework, this paper also introduces \textbf{ATR2-HUTD} dataset, a large-scale UAV-borne HUTD dataset designed to overcome the limitations of existing datasets.  
ATR2-HUTD comprises three sub-datasets—ATR2-HUTD-Lake, ATR2-HUTD-River, and ATR2-HUTD-Sea—collected from LiuYang, Changsha, and Sanya, China, respectively.  
By encompassing diverse lacustrine, riverine, and coastal environments with varying water conditions, ATR2-HUTD mitigates the scene diversity limitations in current datasets, improving model generalization to real-world aquatic settings.  
Additionally, ATR2-HUTD features larger image dimensions, ranging from $2304 \times 640$ to $3536 \times 640$ pixels, providing enhanced spatial details and greater background variability.  
This expanded scale surpasses most existing datasets, alleviating data size constraints and enriching the foundation for model training and evaluation.  
As summarized in Tab.~\ref{dataset}, ATR2-HUTD introduces greater realism and complexity, establishing a more rigorous benchmark for advancing robust and generalizable HUTD methodologies.  
\par

The rest of this paper is structured as follows:  
Section~\ref{sec:3} details the ATR2-HUTD dataset.  
Section~\ref{sec:4} introduces and analyzes the proposed UTD framework.  
Section~\ref{sec:5} presents the experimental results and analysis.  
Section~\ref{sec:6} concludes the paper and discusses future research directions.  

\section{Proposed Dataset}\label{sec:3} 
\subsection{Study region and hardware}\label{sec:2.1}
In this subsection, we introduce the study regions and data acquisition hardware.
\par
\textbf{(1) Study Regions.}  
To investigate the nearshore HUTD problem, three regions with distinct hydrological and environmental characteristics were selected.  
\par
\begin{table}[!t]
    \centering
    \footnotesize
    \renewcommand\arraystretch{1.25}
    \caption{The parameters of the Headwall Nano-Hyperspec sensor.}
    \setlength{\tabcolsep}{6.15mm} 
    {
        \scalebox{0.9}
        { 
    \begin{tabular}{>{\columncolor[HTML]{F0F0F0}}c >{\columncolor[HTML]{FFFFFF}}c}
        \toprule
        \rowcolor[HTML]{C0C0C0}
        \textbf{Parameters} & \textbf{Values} \\    
        \midrule    
        Wavelength range & 400-1000 $nm$ \\    
        Spatial bands & 640 \\    
        Spectral bands & 270 \\  
        Dispersion/pixel & 2.2 $nm$/pixel\\  
        FWHM slit image & 6 $nm$ \\  
        Integrated 2nd order filter & Yes \\  
        Entrance slit width & 20 $\mu m$ \\  
        Bit depth & 12 bit \\
        Detector pixel pitch & 7.4 $\mu m$ \\
        Weight without lens and GPS& $0.5 kg$ \\      
        Size & $7.62 cm \times 7.62 cm \times 8.74 cm $\\  
        Consumption & $\leq$ 13W (9$\sim$24VDC) \\  
        Focal length & $8 mm$ \\
        \bottomrule
    \end{tabular}}
    }
    \label{table:B1}
\end{table}
\begin{figure*}[!t]     
    \centering                         
    \includegraphics[width=2\columnwidth]{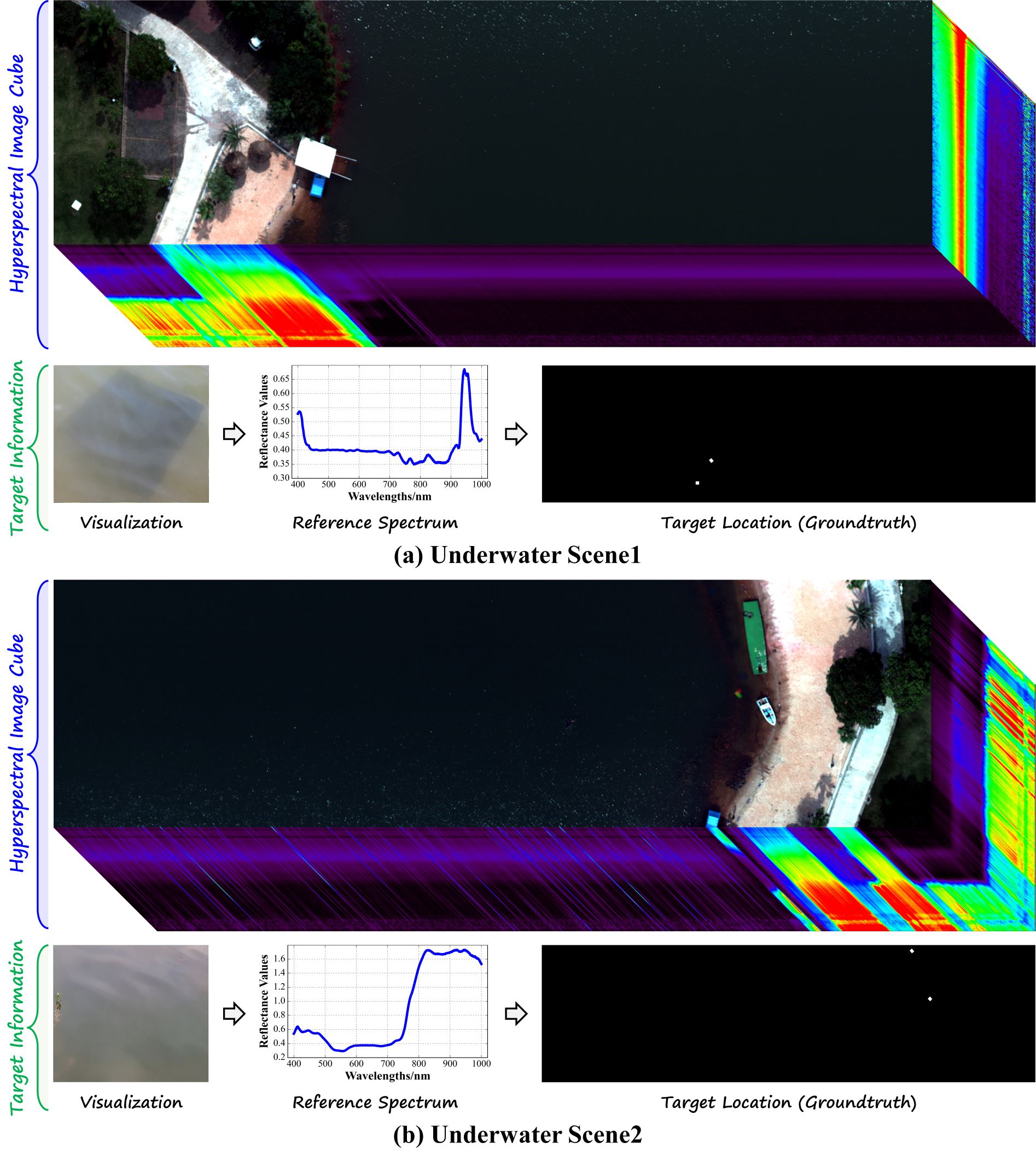}                          
    \caption{The ATR2-HUTD-Lake sub-dataset. (a) Underwater Scene1; (b) Underwater Scene2.}                                  
    \label{fig:B2-1}    
\end{figure*}
\begin{table*}[!t]     
    \centering 
    \footnotesize   
    \renewcommand\arraystretch{1.75}     
    \caption{The crucial information of ATR2-HUTD dataset.}     
    \setlength{\tabcolsep}{2.55mm}    
    {     
        \scalebox{1}
        {
        \begin{tabular}{|cc|c|c|c|c|c|c|}
            \hline
            \multicolumn{2}{|c|}{\textbf{Dataset}}                         & \textbf{Wavelength}         & \textbf{Spectral   Resolution} & \textbf{Image Size}              & \textbf{Spatial   Resolution} & \textbf{Target Type} & \textbf{Target Depth} \\ \hline\hline
            \multicolumn{1}{|c|}{\multirow{2}{*}{\textbf{Lake}}}  & Scene1 & \multirow{6}{*}{400-1000nm} & \multirow{6}{*}{2.2nm}              & \multirow{2}{*}{2304$\times$640 pxiels} & \multirow{2}{*}{5.55   cm}    & Black Metal Plate    & 1.69m, 2.74m          \\ \cline{2-2} \cline{7-8} 
            \multicolumn{1}{|c|}{}                                & Scene2 &                             &                                     &                                  &                               & Blue Metal Plate   & 0.91m, 1.28m          \\ \cline{1-2} \cline{5-8} 
            \multicolumn{1}{|c|}{\multirow{2}{*}{\textbf{River}}} & Scene1 &                             &                                     & \multirow{2}{*}{3536$\times$640 pxiels} & \multirow{2}{*}{4.63 cm}      & Black Plastic Plate  & 1.97m, 1.89m           \\ \cline{2-2} \cline{7-8} 
            \multicolumn{1}{|c|}{}                                & Scene2 &                             &                                     &                                  &                               & Black Metal Plate    & 1.15m, 2.08m                 \\ \cline{1-2} \cline{5-8} 
            \multicolumn{1}{|c|}{\multirow{2}{*}{\textbf{Sea}}}   & Scene1 &                             &                                     & \multirow{2}{*}{3171$\times$640 pxiels} & \multirow{2}{*}{2.78 cm}      & Black Wooden Board   & 0.64m, 1.48m          \\ \cline{2-2} \cline{7-8} 
            \multicolumn{1}{|c|}{}                                & Scene2 &                             &                                     &                                  &                               & Yellow Wooden Board  & 1.35m                 \\ \hline
            \end{tabular} }   
    }
    \label{table:B2}
\end{table*}
The first region, Qianlu Lake in Liuyang City, China, is a mountainous freshwater lake characterized by clear waters, steep terrain, and dense vegetation.  
As a primary freshwater source with low sedimentation and minimal human impact, it provides an optimal setting for UTD studies in low-turbidity freshwater conditions.  
\par
The second region, Xiang River in Changsha, China, is the largest river in the province and a major tributary of Dongting Lake.  
Its high flow rates and substantial sediment transport result in highly turbid waters, particularly during the wet season.  
The riverbed comprises diverse substrates, including silts and sands, creating a complex and dynamic environment for UTD studies in riverine conditions.  
\par
The third region, Yalong Bay in Sanya City, China, represents a coastal marine ecosystem with variable turbidity influenced by coastal currents and biological activity.  
Its seafloor ranges from sandy substrates to coral reefs, with fluctuating salinity and temperature.    
\par
These regions encompass diverse nearshore environments, offering a comprehensive testbed for evaluating UTD methods under varying water conditions and seafloor characteristics.  
\par
\textbf{(2) Hardware.} HSI data for the study regions were collected using a DJI Matrice 300 RTK (M300 RTK) UAV platform, equipped with real-time kinematic (RTK) capabilities.  
With a maximum payload capacity of 9 kg and a flight endurance of up to 55 minutes, the UAV enables extensive data acquisition. The RTK integration ensures centimeter-level positioning accuracy, essential for precise target annotation georeferencing.  
\par
The hyperspectral sensor used is the Headwall Nano-Hyperspec imaging sensor, known for its high spectral resolution and compact design, ideal for UAV-based remote sensing in dynamic nearshore environments.  
Detailed specifications of the sensor are provided in Tab.~\ref{table:B1}, demonstrating its capability to capture a broad spectral range crucial for analyzing complex underwater and nearshore scenes.   
\par
\begin{figure*}[!t]             
    \centering               
    \includegraphics[width=2\columnwidth]{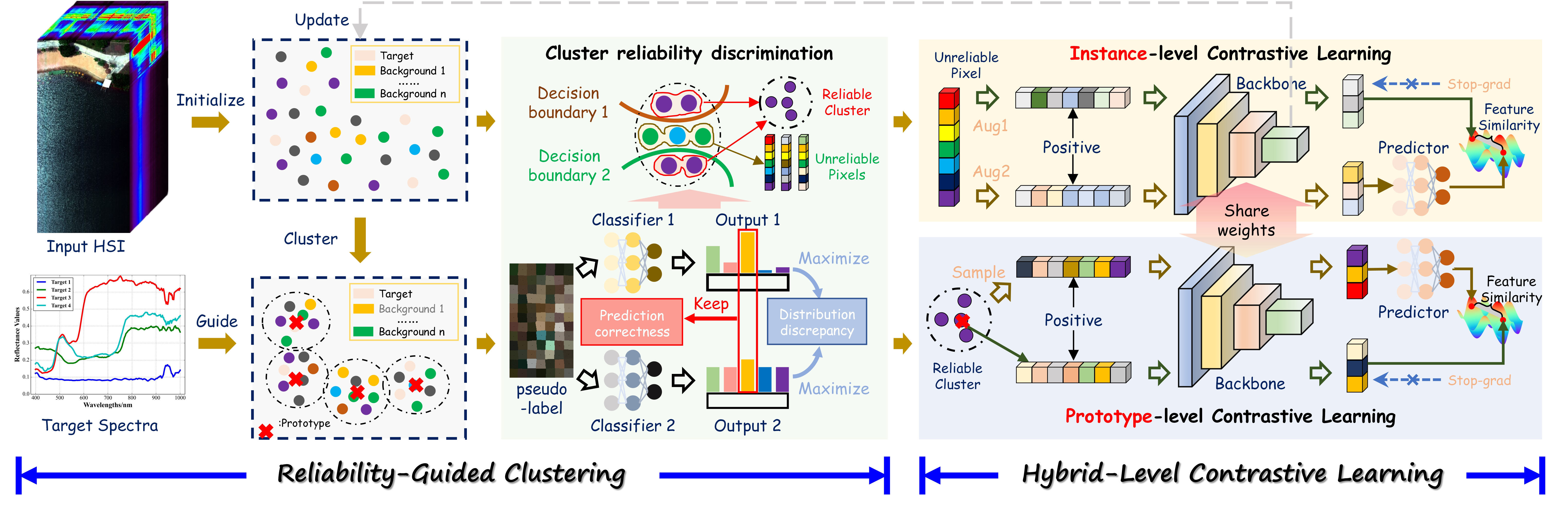}                
    \caption{The flowchart of the proposed underwater target detection framework.}             
    \label{fig:C1}   
\end{figure*}
A field survey utilizing GPS technology was conducted to record the geospatial coordinates of underwater targets, providing accurate ground-truth annotations for the HSI data.  
These annotations are critical for ensuring precise target identification and localization, thereby enhancing the training and evaluation of HUTD models and improving model robustness and performance assessment.
\subsection{ATR2-HUTD dataset}\label{sec:2.2}
This paper introduces the ATR2-HUTD dataset, a novel large-scale benchmark for nearshore UTD, addressing the data scarcity issue while evaluating the proposed method's efficiency and effectiveness.  
The dataset comprises three UAV-borne hyperspectral sub-datasets: ATR2-HUTD-Lake, ATR2-HUTD-River, and ATR2-HUTD-Sea, collected from nearshore regions with diverse water types and underwater targets.  
Key details of these datasets are summarized in Tab.~\ref{table:B2}.
Fig.~\ref{fig:B2-1} illustrates the ATR2-HUTD-Lake sub-dataset as an example, showcasing the underwater scenes and target types.
\par
\textbf{(1) ATR2-HUTD-Lake Sub-dataset:}  
The ATR2-HUTD-Lake sub-dataset was collected on July 6, 2021, between 14:34 and 15:42 at Qianlu Lake, Liuyang City, Hunan Province, China, under clear skies, mild sunlight, and ambient conditions of 25$^\circ$C temperature, 74\% relative humidity, and 1.7 km/h wind speed. Two nearshore regions were surveyed: one with black plastic plates submerged at depths of 1.69 m and 2.74 m, and the other with dark blue plates at depths of 0.91 m and 1.28 m. The UAV operated at 60 m altitude, providing a spatial resolution of 5.55 cm. Hyperspectral images ($2304 \times 640$ pixels) spanned 400-1000 nm with 2.2 nm spectral resolution. Reference spectra were captured on land for target identification. Fig.~\ref{fig:B2-1} presents the dataset overview, including reference spectra and ground truths.
\par
\textbf{(2) ATR2-HUTD-River Sub-dataset:}  
The ATR2-HUTD-River sub-dataset was acquired on July 10, 2024, from 10:27 to 11:09 at Xiang Lake, Changsha City, Hunan Province, China, under clear and sunny conditions with 27$^\circ$C temperature, 78\% humidity, and 2.1 m/s wind speed. Two riverine scenes were surveyed: one with black plastic plates submerged at 1.97 m and 1.89 m, and the other with a black metal plate at 1.15 m. The UAV operated at 50 m altitude, achieving 4.63 cm spatial resolution. Images ($3536 \times 640$ pixels) covered 400-1000 nm with 2.2 nm spectral resolution. Land-based reference spectra were also recorded. 
\par
\textbf{(3) ATR2-HUTD-Sea Sub-dataset:}  
The ATR2-HUTD-Sea sub-dataset was collected on June 5, 2023, between 14:58 and 15:17 at Xiaolong Bay, Sanya City, Hainan Province, China, under clear skies, strong sunlight, 32$^\circ$C temperature, 83\% humidity, and 1.5 m/s wind speed. Two coastal scenes were surveyed: one with black wooden boards submerged at 0.64 m and 1.48 m, and the other with yellow boards at 1.35 m depth. The UAV operated at 30 m altitude, yielding 2.78 cm spatial resolution. Hyperspectral images ($3171 \times 640$ pixels) spanned 400-1000 nm with 2.2 nm resolution. Reference spectra were obtained on land for target identification. 
\par
\section{Methodology}\label{sec:4}
As shown in Figure~\ref{fig:C1}, we propose \textbf{HUCLNet}, a hybrid-level contrastive learning framework aimed at addressing the challenges of nearshore UTD. 
HUCLNet comprises two primary components: the RGC module and the HLCL module.
\par
In the RGC module, the input HSI is decomposed into individual pixels, which are clustered using an unsupervised method. 
A reference spectrum selects a specific cluster, and a novel reliability criterion is introduced to classify clusters into reliable and unreliable instances.
Building on this, the HLCL module leverages contrastive learning to enhance feature discrimination at both the prototype and instance levels. 
\par
HUCLNet operates in two alternating steps: 
(1) Assigning pixel features to clusters and classifying reliable clusters and unreliable instances through a self-paced learning framework (Section \ref{sec3.2}); 
(2) Optimizing spatial-spectral feature extractors using hybrid contrastive learning, progressively updating pixel feature representations through encoded features (Section \ref{sec3.1}).
\par
\subsection{Reliability-Guided Clustering}\label{sec3.2}
The primary objective of HUCLNet is to establish a discriminative feature space where underwater target spectra align with a reference spectrum while remaining distinctly separated from the background through contrastive learning. 
In the absence of supervision to classify pixels in the input HSI as target or background, unsupervised clustering is employed to infer categorical information. 
To address the specific challenges of HUTD, where the reference spectrum is the only prior knowledge, a reference spectrum-guided clustering method is introduced. 
This method incorporates the reference spectrum to enhance clustering reliability, facilitating a more precise distinction between underwater targets and background.

\begin{figure}[!t]                  
    \centering                    
    \includegraphics[width=1\columnwidth]{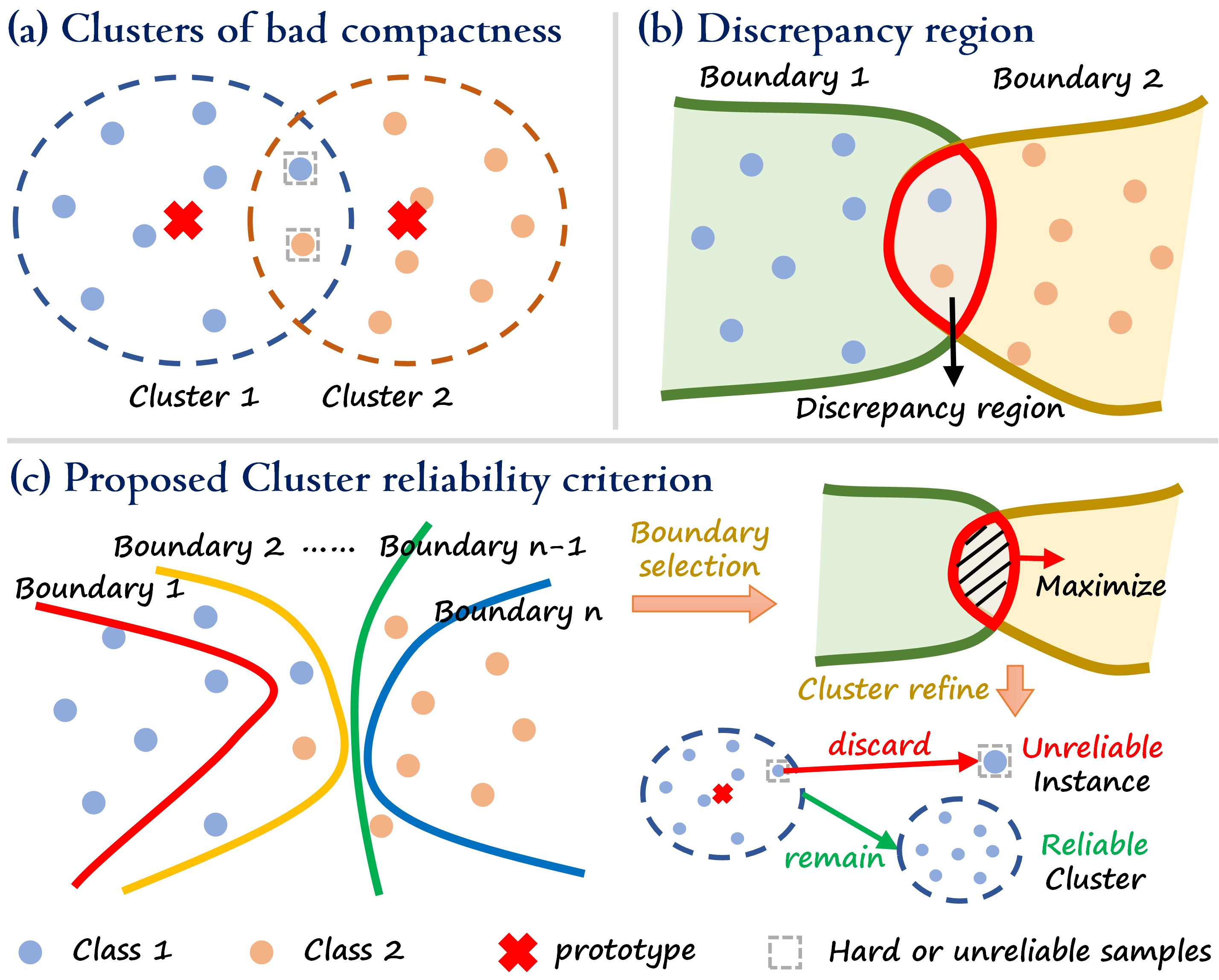}                     
    \caption{The illustration of classifier discrepancy maximum rule.}                  
    \label{fig:C3}    
\end{figure}

Accurate clustering improves HUTD performance by providing the categorical information essential for the hybrid-level contrastive learning module. 
However, as shown in Figure~\ref{fig:C3} (a), clustering often exhibits poor compactness, particularly in early training stages when feature representations lack discriminability. 
This results in clusters containing noisy samples, especially those far from cluster prototypes. 
Incorporating these noisy samples directly into the HLCL module risks degrading its performance and stability. 
To address this, refining clustering results by distinguishing reliable from unreliable clusters is crucial. 
We propose a cluster reliability criterion that evaluates cluster consistency by measuring distances to classifier decision boundaries, based on classifier discrepancy maximization.

Based on these insights, we propose the RGC module, which integrates three key components: the reference spectrum-guided clustering method, the classifier discrepancy maximization rule, and the cluster reliability criterion.

\subsubsection{Reference Spectrum-based Clustering Method}\label{sec3.2.1}
In the nearshore HUTD task, the target regions are limited in spatial extent compared to the extensive background, resulting in significant data imbalance, with far fewer target pixels than background pixels. 
This imbalance poses challenges for model training and generalization. 
However, nearshore areas with shallow waters and varied seabed types exhibit a wide range of spectral characteristics in the background. 
These variations enable the background to be subdivided into multiple distinct types, transforming the \emph{binary target-background} classification problem into a \emph{multi-class target-background} classification problem.

A straightforward method to categorize training samples without supervision is to apply an unsupervised clustering algorithm, such as K-Means~\cite{Sinaga2020}. 
However, conventional clustering algorithms fail to leverage the reference spectrum as prior knowledge. 
To address this, we propose incorporating the reference spectrum as a fixed prototype in the clustering process. 
Building on the DeepCluster approach~\cite{caron2018deep}, we introduce the Reference Spectrum-based Clustering (RSC) method, which explicitly integrates the reference spectrum to improve clustering accuracy and reliability.
\par 
Given a training sample set $\mathbb{X} = \{\bm{x}_1, \bm{x}_2, \cdots, \bm{x}_n\}$, the RSC method uses the corresponding transformed features $\{\bm{h}_1, \bm{h}_2, \cdots, \bm{h}_n\}$ as inputs, obtained through a transformation function $G(\cdot)$, where $\bm{h}_i = G(\bm{x}_i)$. 
The transformation function $G(\cdot)$ is defined as follows, depending on the training epoch:
\begin{equation}\label{eq:transformation}
    G(\bm{x}_i) = 
    \begin{cases}
        \mathcal{I} (\bm{x}_i), & \text{if epoch} = 1, \\
        F(\bm{x}_i|\Theta), & \text{if epoch} > 1,
    \end{cases}
\end{equation}
where $\mathcal{I} (\cdot)$ is the identity mapping function, and $F(\cdot|\Theta)$ represents the backbone network within the hybrid contrastive learning framework. 
In the first training epoch, the identity mapping function $\mathcal{I} (\cdot)$ is used because the feature extraction network is randomly initialized and cannot effectively extract spectral feature vectors at this stage. 
Thus, raw spectral features are employed for unsupervised clustering to avoid negative impacts from the underdeveloped network.
\par
The goal of The RSC method partitions the samples into $k+1$ distinct groups based on an assignment criterion. Specifically, it determines the prototype matrix $\mathbb{P} = \{\bm{p}_1, \bm{p}_2, \cdots, \bm{p}_k\}$ and cluster assignments $\bm{a}_i = \{a_1, a_2, \cdots, a_{k+1}\}$ for each sample $\bm{x}_i$ by solving the following optimization problem:
\begin{equation}
    \begin{gathered} \label{cluster}
        \min_{\mathbb{P}} \frac{1}{n} \sum_{i=1}^n \min_{\bm{a}_i \in \{0,1\}^{k+1}} \left\| \bm{h}_i - \{\mathbb{P}, \text{stopgrad}(\bm{h}_{\text{ref}})\} \cdot \bm{a}_i \right\|_2^2, \\
        \text{s.t.} \quad \bm{a}_i^{\top} \bm{1}_{k+1} = 1,
    \end{gathered}
\end{equation}
where $\bm{1}_{k+1}$ is a vector of ones of dimension $k+1$, and $\bm{h}_{\text{ref}} = G(\bm{x}_{\text{ref}})$ represents the transformed feature of the reference spectrum $\bm{x}_{\text{ref}}$. The term $\text{stopgrad}(\bm{h}_{\text{ref}})$ ensures the reference spectrum's transformed feature is treated as a fixed constant during training. To prevent trivial solutions, such as assigning all samples to a single cluster, an equipartition constraint~\cite{caron2020unsupervised} is applied.

\subsubsection{Classifier Discrepancy Maximization Rule}\label{sec:3.2.2}
Unreliable samples are typically located near decision boundaries~\cite{Hearst1998}. As shown in Figure~\ref{fig:C3} (b), these can be identified where decision boundaries of two classifiers, trained on the same dataset, exhibit discrepancies. Larger discrepancy regions help identify a broader set of unreliable samples, in line with the principle of "preferring quality over quantity." To refine clustering results, we propose a classifier discrepancy maximization rule, illustrated in Figure~\ref{fig:C3} (c), which maximizes the cross-entropy between the outputs of two classifiers while maintaining a consistency constraint to improve classification reliability.

The pseudo-labels $\mathbb{Y} = \{y_1, y_2, \dots, y_n\}$ for the training samples are derived from the clustering results, where $y_i = \arg\max_{i \in \{1, 2, \dots, k+1\}} a_i$. Considering two classifiers, $\bm{C}_1$ and $\bm{C}_2$, sharing the same architecture but with different initial weights, the probabilistic outputs for sample $\bm{x}_i$ are $\bm{q}_1^i \triangleq \bm{C}_1(\bm{x}_i)$ and $\bm{q}_2^i \triangleq \bm{C}_2(\bm{x}_i)$. To maximize decision boundary discrepancies between $\bm{C}_1$ and $\bm{C}_2$, we maximize the cross-entropy between their outputs, $\{\bm{C}_j(\bm{x}_i)\}_{j=1}^2$, while applying a prediction consistency constraint to prevent misclassification of reliable samples. The classifier discrepancy maximization rule is formulated as:
\begin{multline}\label{rule}
    \arg\max_{\bm{C}_1, \bm{C}_2} \sum_{i = 1}^{n} \Big(CE(\bm{q}_1^i, \bm{q}_2^i) + CE(\bm{q}_2^i, \bm{q}_1^i) \\ - CE(\mathbb{Y}, \bm{q}_1^i) - CE(\mathbb{Y}, \bm{q}_2^i)\Big),
\end{multline}
where $CE(\bm{m}, \bm{n})$ denotes the cross-entropy between distributions $\bm{m}$ and $\bm{n}$, defined as:
\begin{equation}
    CE(\bm{m}, \bm{n}) = \sum_{k = 1}^{K} m_k \log(n_k),
\end{equation}
with $K$ being the dimension of the distribution vector, and $m_k$ and $n_k$ representing the $k$-th elements of $\bm{m}$ and $\bm{n}$, respectively.
\subsubsection{Cluster Refinement Strategy}
As outlined in Section \ref{sec:3.2.2}, samples near the decision boundaries of two classifiers are prone to unreliability. Therefore, classifier disagreement, as specified by the rule in Eq. (\ref{rule}), serves as a mechanism to identify unreliable samples within a cluster. The reliability of a training sample $\bm{x}_i$ is quantified using the following criterion:
\begin{equation}\label{Cluster Reliability Criterion}
    \varrho_{i} = 
    \begin{cases}
        1, & \text{if } [\bm{C}_1(\bm{x}_i)]_{\max} = [\bm{C}_2(\bm{x}_i)]_{\max}, \\        
        0, & \text{if } [\bm{C}_1(\bm{x}_i)]_{\max} \neq [\bm{C}_2(\bm{x}_i)]_{\max}. 
    \end{cases}
\end{equation}
Here, $[\cdot]_{\max}$ denotes the index of the maximum value in the vector, and $\varrho_{i}$ serves as the reliability indicator. A value of $\varrho_{i} = 1$ indicates that the sample $\bm{x}_i$ is reliable. This criterion refines the clustering outcomes from Eq. (\ref{cluster}), partitioning the training samples $\mathbb{X}$ into reliable clusters $\mathbb{X}_c = \{\mathbb{X}_{c_1}, \dots, \mathbb{X}_{c_t}\}$, where $t \leq k+1$ is the number of reliable clusters, and unreliable instances $\mathbb{X}_u$. Consequently, $\mathbb{X}_c \cup \mathbb{X}_u = \mathbb{X}$.
\par
\subsection{Hybrid-Level Contrastive Learning}\label{sec3.1}
As discussed in Section~\ref{sec:1}, nearshore Hyperspectral Underwater Target Detection (HUTD) faces two main challenges: \emph{dependence on bathymetric models} and \emph{limited target characterization}. 
To address these challenges, we introduce a hybrid-level contrastive learning framework that integrates instance-level and prototype-level contrastive learning modules. 
Data augmentation is a critical aspect of contrastive learning, but conventional methods, which primarily focus on spatial transformations, often compromise the spectral integrity of hyperspectral data. 
To counter this, we propose a hyperspectral-specific data augmentation strategy that incorporates unsupervised adversarial training, ensuring the effective preservation and utilization of both spatial and spectral information.

The architecture of the proposed framework is illustrated in Figure~\ref{fig:C1}. Each module comprises two branches: a shared backbone network, $F(\cdot|\Theta_{x})$, and a projection MLP head, $h_{x}(\cdot|\bm{W}_x)$, where $x = \{I, C\}$ denotes the instance-level or prototype-level module. 
To capture coarse-to-fine contrastive semantic information, the backbone networks are shared across modules, \emph{i.e.}, $\Theta_{I} = \Theta_{C}$. For simplicity, we denote all backbone networks as $F(\cdot|\Theta)$ throughout the article, implemented using 3D-ResNet50~\cite{Jiang2019} for feature extraction. The projection MLP head predicts one view based on the output of the other, facilitating contrastive learning. Both modules adopt identical projection MLP structures~\cite{ChenH21}, without weight sharing.

\subsubsection{Instance-Level Contrastive Learning}\label{sec3.1.1}
The instance-level contrastive learning module addresses the challenge of unreliable instances, typically distant from cluster prototypes due to their poor discriminability.  
It treats each unreliable instance as a separate class, enhancing the model's ability to capture explicit similarities and differences among instances~\cite{Wu2018}.  
This facilitates the extraction of more discriminative feature representations, ensuring sufficient discriminability for the unsupervised clustering strategy in Section~\ref{sec3.2.1}.  
Additionally, this module strengthens the model's ability to characterize targets, providing a robust semantic foundation for the subsequent prototype-level contrastive learning stage.
Let $\bm{x}_{u}^{i} \in \mathbb{X}_{u}$ represent an example, where two augmented views $\hat{\bm{x}}_{u}^{i}$ and $\tilde{\bm{x}}_{u}^{i}$ are generated using the augmentation strategy (see Section \ref{sec3.1.3}). These views are passed through distinct branches, producing output vectors $\bm{\hat{\bm{p}}}^{i}_{u} \triangleq h_{I}(F(\hat{\bm{x}}_{u}^{i}|\Theta)|\bm{W}_I)$ and $\tilde{\bm{z}}_{u}^{i} \triangleq F(\tilde{\bm{x}}_{u}^{i}|\Theta)$. The view prediction error is measured by negative cosine similarity, given by:
\begin{equation}\label{eq:1}
    \mathcal{D}\left(\bm{\hat{\bm{p}}}^{i}_{u}, \tilde{\bm{z}}_{u}^{i}\right) = -\frac{\bm{\hat{\bm{p}}}^{i}_{u}}{\left\|\bm{\hat{\bm{p}}}^{i}_{u}\right\|_2} \cdot \frac{\tilde{\bm{z}}_{u}^{i}}{\left\|\tilde{\bm{z}}_{u}^{i}\right\|_2},
\end{equation}
where $\|\cdot\|_2$ denotes the $\ell_2$-norm. The objective function for the instance-level contrastive learning module is a symmetrized loss:
\begin{equation}\label{eq:2}
    \mathcal{L}_{\text{instance}} = \frac{1}{2} \mathcal{D}\left(\bm{\hat{\bm{p}}}^{i}_{u}, \tilde{\bm{z}}_{u}^{i}\right) + \frac{1}{2} \mathcal{D}\left(\bm{\tilde{\bm{p}}}^{i}_{u}, \hat{\bm{z}}_{u}^{i}\right).
\end{equation}
To prevent model collapse, a stop-gradient operation is applied, reformulating Eq. (\ref{eq:1}) as:
\begin{equation}\label{eq:3}     
    \mathcal{D}\left(\bm{\hat{\bm{p}}}^{i}_{u}, \text{stopgrad}(\tilde{\bm{z}}_{u}^{i})\right),
\end{equation}
where $\text{stopgrad}(\cdot)$ acts as a constant during optimization. The updated learning objective is:
\begin{equation}\label{eq:4}
    \mathcal{L}_{\text{instance}} = \frac{1}{2} \mathcal{D}\left(\bm{\hat{\bm{p}}}^{i}_{u}, \text{stopgrad}(\tilde{\bm{z}}_{u}^{i})\right) + \frac{1}{2} \mathcal{D}\left(\bm{\hat{\bm{z}}}^{i}_{u}, \text{stopgrad}(\tilde{\bm{p}}_{u}^{i})\right).
\end{equation}

\subsubsection{Prototype-Level Contrastive Learning}\label{sec3.1.2}
The prototype-level contrastive learning module enhances the semantic consistency of target representations by aligning homogeneous samples within reliable clusters, thereby improving the separation between target and background clusters.  
Let $\bm{x}_{c_m}^{i}$ denote the $i$-th pixel within the $m$-th reliable cluster $\mathbb{X}_{c_m}$, with $\bar{\bm{x}}_{c_m}$ as its corresponding prototype. These samples are processed through separate branches, yielding output vectors $\bm{p}^{i}_{c_{m}} \triangleq h_{C}(F(\bm{x}^{i}_{c_m}|\Theta)|\bm{W}_C)$ and $\bar{\bm{z}}_{c_{m}} \triangleq F(\bar{\bm{x}}_{c_m}|\Theta)$.

In contrast to instance-level learning, the prototype $\bar{\bm{x}}_{c_m}$ serves as a stationary reference during the alignment process to ensure stable convergence. The view prediction loss is given by:
\begin{equation}\label{eq:5}   
    \mathcal{D}\left(\bm{p}^{i}_{c_{m}}, \bar{\bm{z}}_{c_{m}}\right)=-\frac{\bm{p}^{i}_{c_{m}}}{\left\|\bm{p}^{i}_{c_{m}}\right\|_2} \cdot \frac{\bar{\bm{z}}_{c_{m}}}{\left\|\bar{\bm{z}}_{c_{m}}\right\|_2},
\end{equation}
with the loss for cluster pre-alignment defined as:
\begin{equation}\label{eq:6} 
    \mathcal{L}_{cluster}^{pre} = \mathcal{D}\left(\bm{p}^{i}_{c_{m}}, \text{stopgrad}(\bar{\bm{z}}_{c_{m}})\right).
\end{equation}
To enhance inter-cluster discrepancy, we apply the InfoNCE loss to the prototypes $\{\bar{\bm{x}}_{c_1}, \dots, \bar{\bm{x}}_{c_t}\}$:
\begin{equation}\label{eq:7} 
    \resizebox{0.89\hsize}{!}{$\mathcal{L}_{cluster}^{pro}=-\sum_{i=1}^t\log \frac{\exp (\bar{\bm{x}}_{c_i} \cdot \bar{\bm{x}}_{c_i}^{+} / \tau)} {\sum_{j=1}^t \left(\exp (\bar{\bm{x}}_{c_i} \cdot \bar{\bm{x}}_{c_j} / \tau) + \exp (\bar{\bm{x}}_{c_j} \cdot \bar{\bm{x}}_{c_j}^{+} / \tau)\right)}.$}
\end{equation}
\par
The overall objective for the prototype-level contrastive learning module is the linear combination:
\begin{equation}\label{eq:8} 
    \mathcal{L}_{cluster} = \mathcal{L}_{cluster}^{pre} + \mathcal{L}_{cluster}^{pro}.
\end{equation}
\begin{figure}[!ht]                  
    \centering                    
    \includegraphics[width=1\columnwidth]{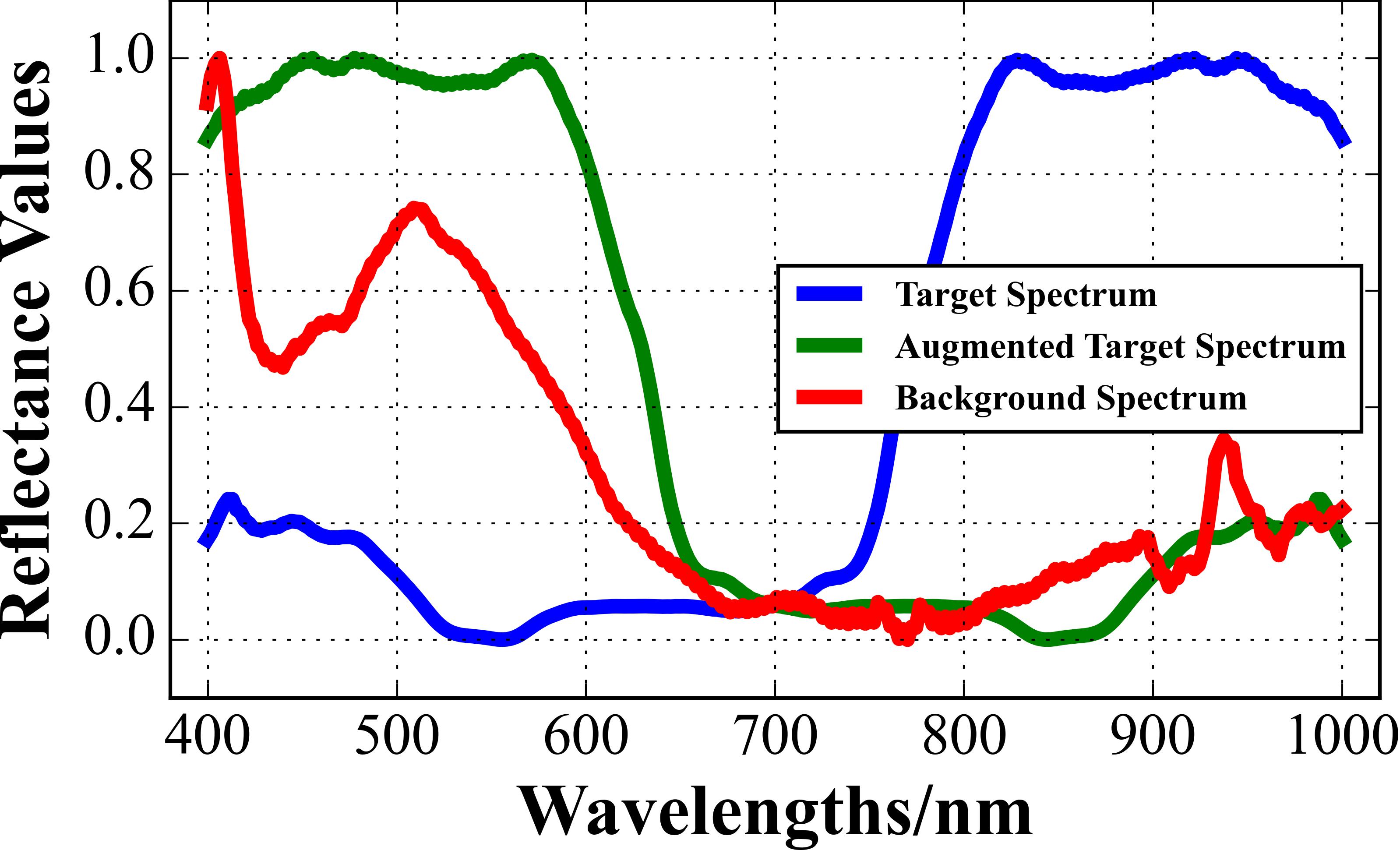}                     
    \caption{Illustration of spectral distortion induced by the flipping augmentation operation. The augmented target spectrum exhibits minimal deviation from the background spectrum, yet demonstrates a pronounced discrepancy from the original target spectrum.}              
    \label{fig:aug}    
\end{figure}
\subsubsection{Hyperspectral-Oriented Data Augmentation}\label{sec3.1.3}
\begin{figure*}[!t]             
    \centering               
    \includegraphics[width=2\columnwidth]{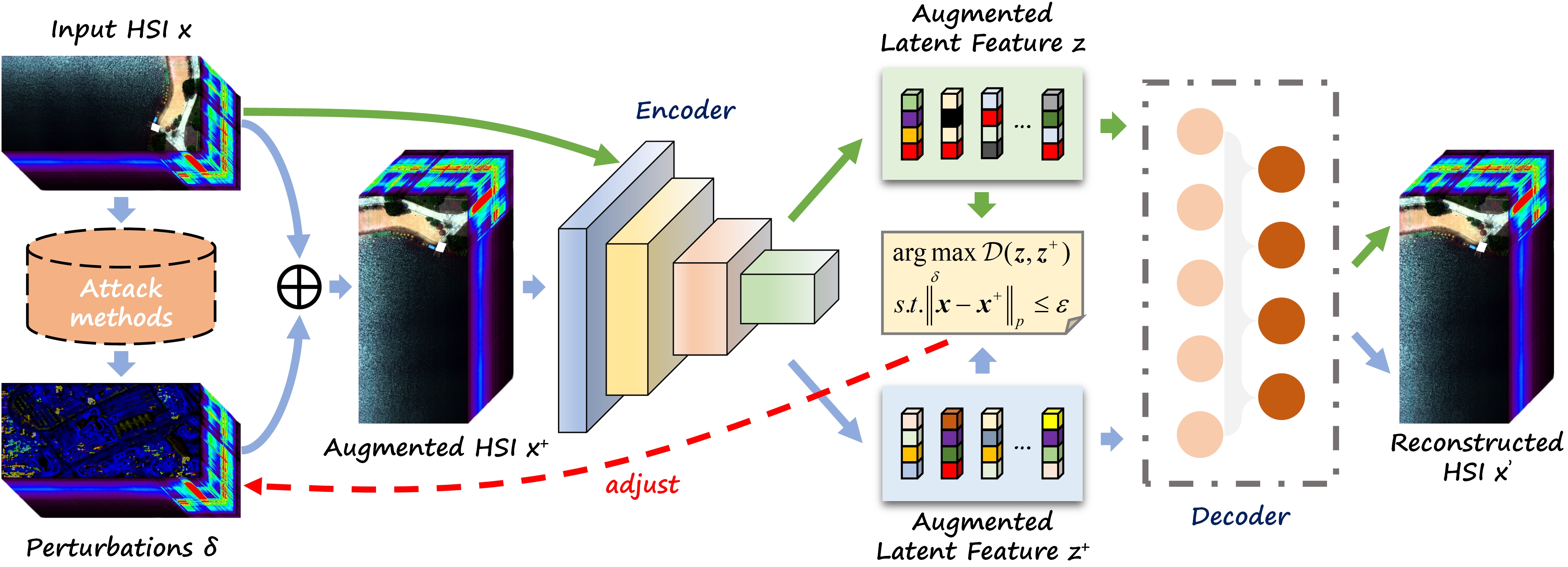}                
    \caption{The process of using adversarial training for hyperspectral data augmentation.}             
    \label{fig:C2}   
\end{figure*}

Traditional data augmentation techniques, such as flipping, rotation, and masking, can compromise spectral integrity~\cite{Liu12024}. For instance, flipping along the spectral dimension alters the spectral semantics, potentially transforming a target spectrum into a background one, as shown in Figure~\ref{fig:aug}. 
To address this, we propose a hyperspectral-specific data augmentation strategy, illustrated in Figure~\ref{fig:C2}. This strategy includes two stages: unsupervised pretraining and self-supervised adversarial training.

\textbf{Unsupervised Pretraining.} To ensure semantic consistency between the augmented and original samples, we introduce an encoder-decoder network. 
The encoder $F(\cdot|\Theta)$ extracts a discriminative representation $\bm{z}$, and the decoder $g(\cdot|\bm{W})$ reconstructs the original pixel. The optimization objective for the unsupervised pretraining is:
\begin{equation}\label{eq:8-1}
    \Theta^{\ast}, \bm{W}^{\ast} = \arg\min\limits_{\Theta, \bm{W}} ||g(\bm{z}|\bm{W}) - \bm{x}||_2.
\end{equation}
For augmented samples, the reconstruction error is minimized as:
\begin{equation}\label{eq:10}
    \arg\min\limits_{\bm{\delta}} ||g(F(\bm{x}^{+}|\Theta)|\bm{W}) - \bm{x}||_2.
\end{equation}
This ensures that spectral semantics are preserved during augmentation, enhancing the quality of contrastive learning for hyperspectral data.
\par
\textbf{Self-Supervised Adversarial Training.} 
Adversarial training effectively generates augmented samples by applying controlled perturbations to the original data. In this work, we incorporate an adversarial training stage to generate the perturbation $\bm{\delta}$. Unlike traditional adversarial training, which relies on downstream task outputs, we guide perturbation generation through latent feature discrepancies in a self-supervised manner.

The encoder $F(\cdot|\Theta)$, with pretrained weights $\Theta^{\ast}$, computes latent representations for the original pixel $\bm{x}$ and its perturbed version $\bm{x}^+$ as $\bm{z}_{\ast} \triangleq F(\bm{x}|\Theta^{\ast})$ and $\bm{z}_{\ast}^+ \triangleq F(\bm{x}^+|\Theta^{\ast})$. To minimize spectral distortion, the perturbation is constrained by an $\ell_p$-norm with an upper bound $\epsilon$, \emph{i.e.}, $||\bm{\delta}||_p \leq \epsilon$. Semantic consistency is further ensured by incorporating the constraint in Eq.~(\ref{eq:8-1}). The adversarial training objective is then formulated as:
\begin{equation}\label{eq:9}        
    \begin{gathered}      
        \bm{\delta}^{\ast} = \arg\max\limits_{\bm{\delta}} \left( ||\bm{z}^+_{\ast} - \bm{z}_{\ast}||_2 - ||g(\bm{z}^+_{\ast}|\bm{W}^{\ast}) - \bm{x}||_2 \right), \\
        \text{s.t.} \ ||\bm{\delta}||_p \leq \epsilon,
    \end{gathered}   
\end{equation}
where $||\cdot||_p$ denotes the $\ell_p$-norm, and $p$ corresponds to the attack method used. The optimization problem is solved using established attack algorithms, such as FGSM~\cite{GoodfellowSS14}, PGD~\cite{MadryMSTV18}, and FAB~\cite{Croce020}, enabling diverse data augmentations.
\par
\subsection{Self-Paced Learning Paradigm}\label{sec3.3}
The clustering results from Section~\ref{sec3.2} guide the HLCL module in Section~\ref{sec3.1}, facilitating more accurate target characterization. 
Simultaneously, the refined target characterization improves clustering performance, even under unsupervised conditions, establishing a mutually reinforcing relationship between clustering and target characterization. However, during early training, both clustering and target characterization are unreliable. Incorporating erroneous clustering information into the HLCL module or performing clustering with incomplete target representations may lead to error propagation, diminishing performance and stability of HUCLNet. To address this, we propose a Self-Paced Learning (SPL) paradigm that progressively improves clustering reliability as spatial-spectral feature extractors become more robust, stabilizing training and enhancing model performance.

The SPL paradigm alternates between the HLCL and RGC modules in a self-paced manner. Initially, the RGC module partitions the training samples into unreliable instances and reliable clusters, as detailed in Section~\ref{sec3.2}. According to the self-paced learning principle, the reliable clusters are treated as easy samples to enhance target characterization, while the unreliable instances serve as hard samples to refine pixel-level discriminability. Once the reliable clusters and unreliable instances are identified, they are passed into the HLCL module. The HLCL module then refines the spatial-spectral feature representations of the training samples, as explained in Section~\ref{sec3.1}. With these updated representations, the RGC module is invoked again to reassign the samples into reliable clusters and unreliable instances. This refined clustering outcome improves the feature learning capacity of the HLCL module, leading to more discriminative feature representations. These refined features, in turn, enable more accurate clustering results. The learning process alternates between the HLCL and target spectrum-guided clustering methods in a self-paced manner until convergence is achieved.

\subsection{Hyperspectral Underwater Target Detection}\label{sec3.4}
Upon completing the self-paced learning process, the refined spatial-spectral feature representations from the HLCL module can be used for underwater target detection. Let $\Theta^{\ast}$ denote the optimal network weights for the backbone network $F(\cdot|\Theta)$ within the HLCL module. The pixels of the input Hyperspectral Image (HSI) $\mathbb{X} = \{\bm{x}_1, \bm{x}_2, \dots, \bm{x}_n\}$ and the reference spectrum $\bm{x}_{\text{ref}}$ are processed through the backbone network to obtain their corresponding feature representations:
\begin{equation}
    \begin{gathered}
        \bm{Z} = \{F(\bm{x}_1|\Theta^{\ast }), F(\bm{x}_2|\Theta^{\ast }), \dots, F(\bm{x}_n|\Theta^{\ast })\}, \\
        \bm{z}_{\text{ref}} = F(\bm{x}_{\text{ref}}|\Theta^{\ast }).
    \end{gathered}
\end{equation}

The target detection task is reformulated as a pixel-wise similarity measurement between the feature representations $\bm{Z}$ of the input HSI and the reference spectrum $\bm{z}_{\text{ref}}$. This can be achieved using common hyperspectral target detection (HTD) algorithms, such as the Spectral Angle Mapper (SAM)~\cite{KRUSE1993145} and the Constrained Energy Minimization (CEM)~\cite{Manolakis2002}. The final detection results are expressed as:
\begin{equation}
    \bm{d} = \text{Detection}(\bm{Z}, \bm{z}_{\text{ref}}),
\end{equation}
where $\bm{d}$ denotes the detection outcomes for the input HSI, and $\text{Detection}(\cdot)$ refers to the chosen HTD algorithm.
\section{Experiments and analysis} \label{sec:5}
This section presents comprehensive experiments on the ATR2-HUTD dataset to evaluate the effectiveness of the proposed method. 
Section~\ref{sec:4.1} outlines the experimental metrics used. 
Section~\ref{sec:4.2} details the network architecture, comparison methods, experimental setup, and parameter configurations. 
To highlight the superiority of the proposed method, Section~\ref{sec:4.3} provides both quantitative analysis and visual evaluations across all comparison methods. 
Section~\ref{sec:4.4} includes ablation studies to assess the contributions of different model components, while Section~\ref{sec:4.5} presents a parameter sensitivity analysis.
\subsection{Evaluation Indicators}\label{sec:4.1}
To quantitatively assess the performance of the proposed method, we employ three widely recognized evaluation metrics in the HTD field.
\par
\textbf{(\romannumeral1) Receiver Operating Characteristic (ROC)~\cite{ROC, ROC3D}:} 
The ROC curve offers an unbiased, threshold-independent evaluation of detection performance. This paper presents three 2D ROC curves: $( \mathrm{P}_{\mathrm{d}}, \mathrm{P}_{\mathrm{f}})$, $( \mathrm{P}_{\mathrm{d}}, \tau)$, and $( \mathrm{P}_{\mathrm{f}}, \tau)$, along with a 3D ROC curve~\cite{ROC3D} of $(\tau, \mathrm{P}_{\mathrm{d}}, \mathrm{P}_{\mathrm{f}})$ for a comprehensive performance evaluation. A detector with ROC curves closer to the upper left, upper right, and lower left corners generally exhibits superior HTD performance.
\par
\textbf{(\romannumeral2) Area Under the ROC Curve (AUC)~\cite{Zhang2015}:} 
To address challenges in visually comparing ROC curves, we compute the area under each of the three 2D ROC curves: $\text{AUC}_{( \mathrm{P}_{\mathrm{d}}, \mathrm{P}_{\mathrm{f}})}$, $\text{AUC}_{( \mathrm{P}_{\mathrm{d}}, \tau)}$, and $\text{AUC}_{( \mathrm{P}_{\mathrm{f}}, \tau)}$. Larger AUC values indicate better performance, with $\text{AUC}_{( \mathrm{P}_{\mathrm{d}}, \mathrm{P}_{\mathrm{f}})} \to 1$, $\text{AUC}_{( \mathrm{P}_{\mathrm{d}}, \tau)} \to 1$, and $\text{AUC}_{( \mathrm{P}_{\mathrm{f}}, \tau)} \to 0$ signifying superior detection performance. Additionally, two AUC-based metrics are introduced for a more comprehensive evaluation:
\begin{equation}
    \mathrm{AUC}_{\mathrm{OA}} = \mathrm{AUC}_{\left(P_f, P_d\right)} + \mathrm{AUC}_{\left(\tau, P_d\right)} - \mathrm{AUC}_{\left(\tau, P_f\right)},
\end{equation}
\begin{equation}
    \mathrm{AUC}_{\mathrm{SNPR}} = \frac{\mathrm{AUC}_{\left(\tau, P_d\right)}}{\mathrm{AUC}_{\left(\tau, P_f\right)}},
\end{equation}
where higher values of $\mathrm{AUC}_{\mathrm{OA}} \to 2$ and $\mathrm{AUC}_{\mathrm{SNPR}} \to +\infty$ indicate improved detector performance.
\subsection{Experimental Details and Settings}\label{sec:4.2}
\textbf{(\romannumeral1) Experimental Details:} 
The experimental setup and details of the proposed method are as follows. Unless otherwise specified, the parameters are applied consistently across all sub-datasets. The method consists of three core components: the RGC module, the HLCL module, and the SPL strategy, each contributing significantly to performance.

In the RGC module, unsupervised clustering is performed using the K-Means~\cite{Sinaga2020} algorithm, with cluster numbers set to 36, 39, and 42 for the lake, river, and sea sub-datasets, respectively, based on environmental complexity and waterbed characteristics.

The HLCL module employs the 3D-ResNet50~\cite{Jiang2019} network for spectral-spatial feature extraction. To enhance robustness and contrastive learning, untargeted FGSM~\cite{GoodfellowSS14} data augmentation is applied with a maximum perturbation of $\epsilon=0.1$ under the $l_{\infty}$ norm. The hybrid-level contrastive learning framework is trained for 50 epochs per SPL iteration. The Adam optimizer is used with a batch size of 256. The initial learning rate is $5\times10^{-3}$, decaying to $5\times10^{-5}$ through a cosine annealing schedule after 100 epochs, and a weight decay of $1\times10^{-4}$ is applied to reduce overfitting.

The SPL strategy is executed for 10 iterations across all sub-datasets to ensure convergence and computational efficiency.

For HUTD, as described in Section~\ref{sec3.4}, we use learned representations combined with basic hyperspectral detectors. To isolate the effect of detectors on performance, we employ two classic detectors, CEM~\cite{KRUSE1993145} and SAM~\cite{Manolakis2002}, as baseline methods.

\textbf{(\romannumeral2) Experimental Settings:} 
We compare the proposed method against several state-of-the-art (SOTA) HTD and HUTD methods, including two traditional HTD detectors (CEM and SAM), two advanced HTD methods (IEEPST~\cite{IEEPST} and MCLT~\cite{Wang2024}), and four HUTD methods (UTD-Net~\cite{Qi2021}, TUTDF~\cite{LiZheyong2023}, TDSS-UTD~\cite{Li2023}, and NUN-UTD~\cite{Liu2024}).

To ensure fairness, each method is executed with the original hyperparameter settings as specified in their respective publications. All experiments are conducted on a machine equipped with seven NVIDIA A6000 GPUs, an AMD Ryzen 5995WX CPU, and 128 GB of RAM, running Ubuntu 22.04.

\subsection{Main Results} \label{sec:4.3}
\begin{figure*}[!t]                 
    \centering                    
    \includegraphics[width=2\columnwidth]{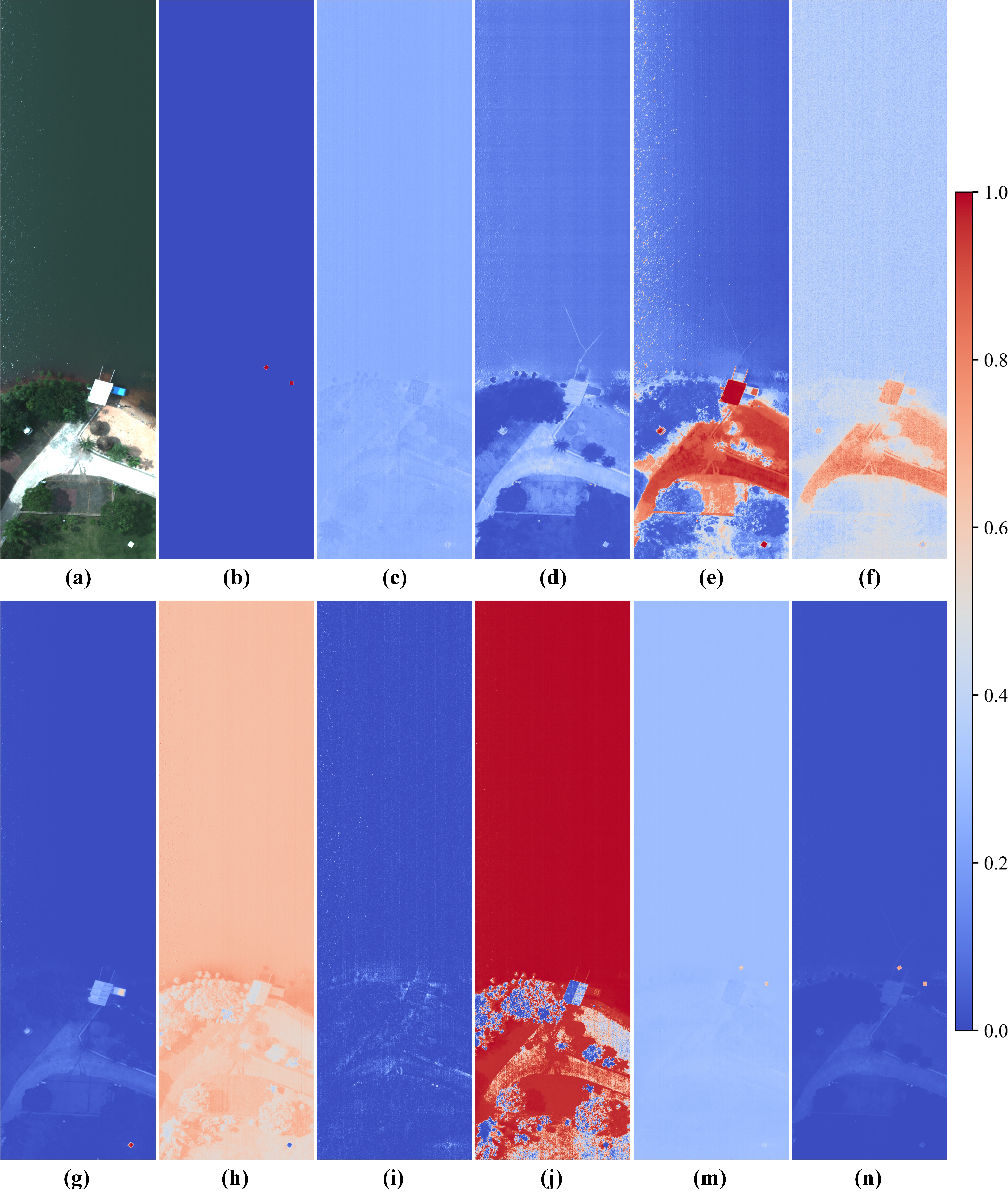}                     
    \caption{Detection maps of ATR2-HUTD Lake Scene1. (a) Pseudo-color image. (b) Ground truth. (c) CEM. (d) SAM. (e) IEEPST. (f) MCLT. (g) UTD-Net. (h) TUTDF. (i) TDSS-UTD. (j) NUN-UTD. (m) HUCLNet+CEM. (n) HUCLNet+SAM.}                  
    \label{fig:C1-1}    
\end{figure*}

\begin{figure*}[!t]                 
    \centering                    
    \includegraphics[width=2\columnwidth]{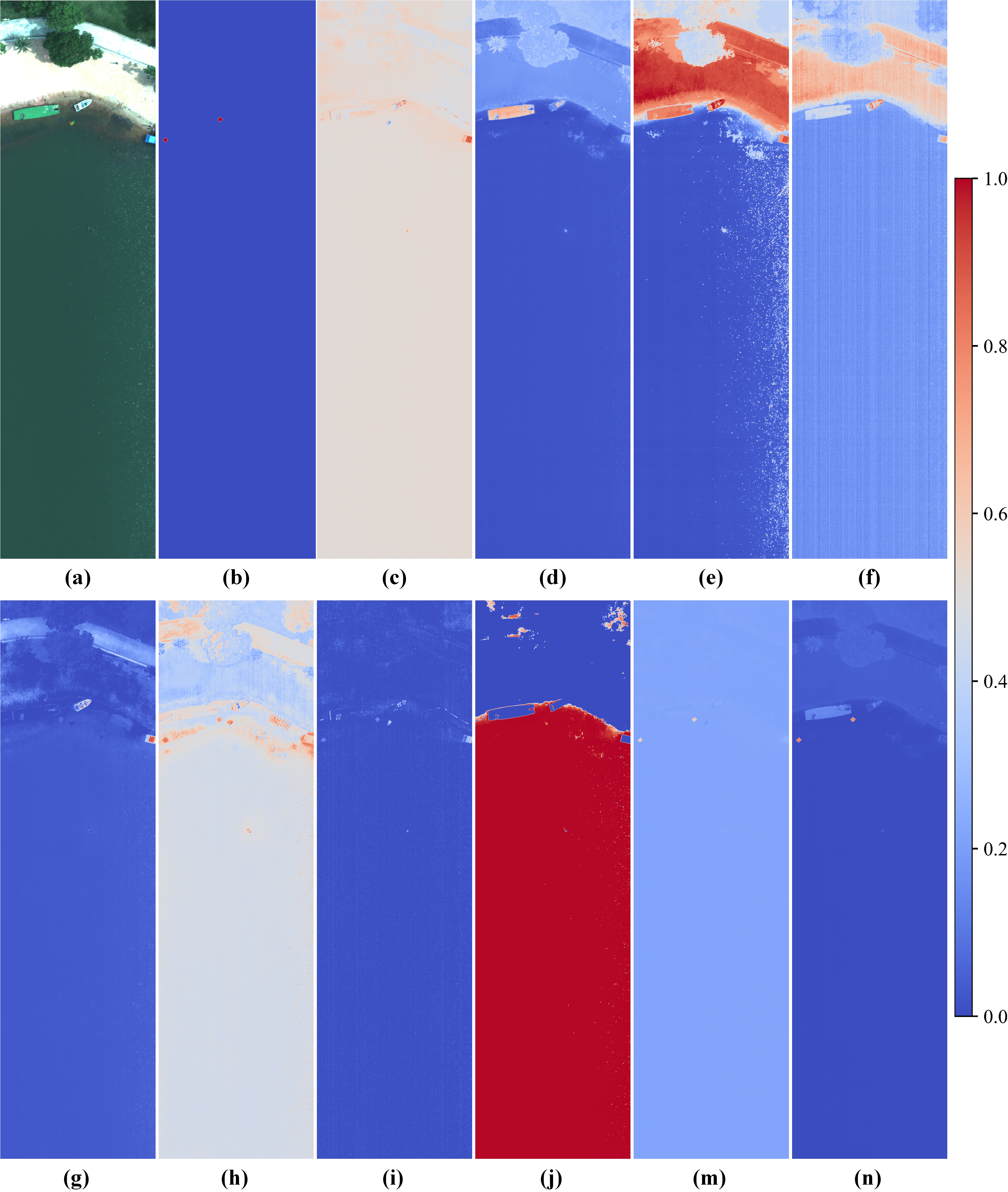}                     
    \caption{Detection maps of ATR2-HUTD Lake Scene2. (a) Pseudo-color image. (b) Ground truth. (c) CEM. (d) SAM. (e) IEEPST. (f) MCLT. (g) UTD-Net. (h) TUTDF. (i) TDSS-UTD. (j) NUN-UTD. (m) HUCLNet+CEM. (n) HUCLNet+SAM.}                    
    \label{fig:C1-2}    
\end{figure*}

\textbf{(\romannumeral1) Detection Maps:} Figs.~\ref{fig:C1-1} to~\ref{fig:C1-2} present detection maps from the ATR2-HUTD-Lake sub-dataset, offering a qualitative comparison of the evaluated methods.
The detection maps of other sub-datasets are provided in the supplementary material.
\par
Traditional methods, such as CEM and SAM, exhibit significant limitations in underwater environments. CEM struggles with background noise suppression, resulting in false positives, while SAM fails to delineate target boundaries and often misses targets, especially in complex scenarios like the ATR2-HUTD River dataset. Its sensitivity to spectral noise and limited adaptability to spectral variations lead to incomplete detection and poor target-background separation.
\par
\begin{figure*}[!t]                 
    \centering                    
    \includegraphics[width=2\columnwidth]{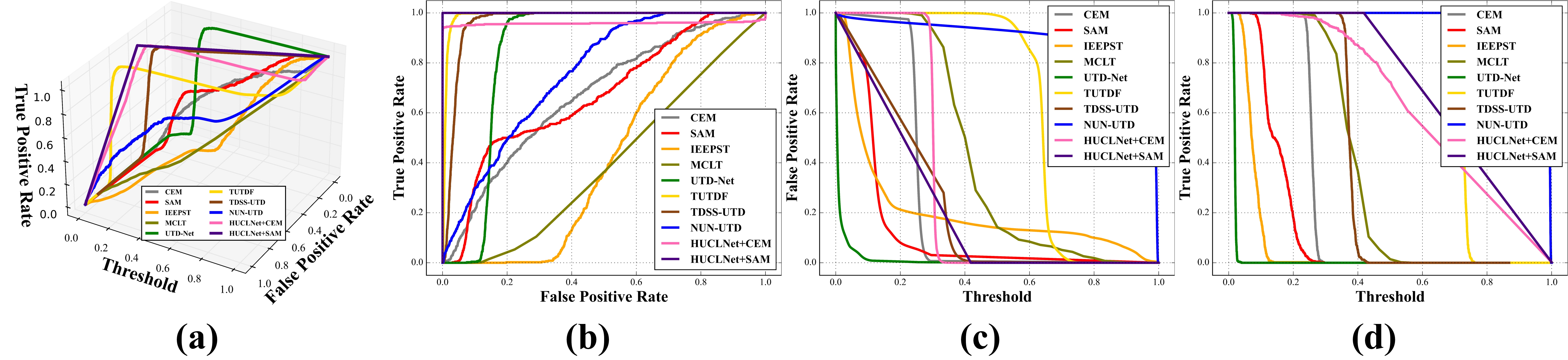}                     
    \caption{ROC curves comparison on ATR2-HUTD Lake Scene1. (a) 3-D ROC curve. (b) 2-D ROC curve of $(P_d, P_f)$. (c) 2-D ROC curve of $(P_f, \tau)$. (d) 2-D ROC curve of $(P_d, \tau)$.}                 
    \label{fig:C2-1}    
\end{figure*}
\begin{figure*}[!t]                 
    \centering                    
    \includegraphics[width=2\columnwidth]{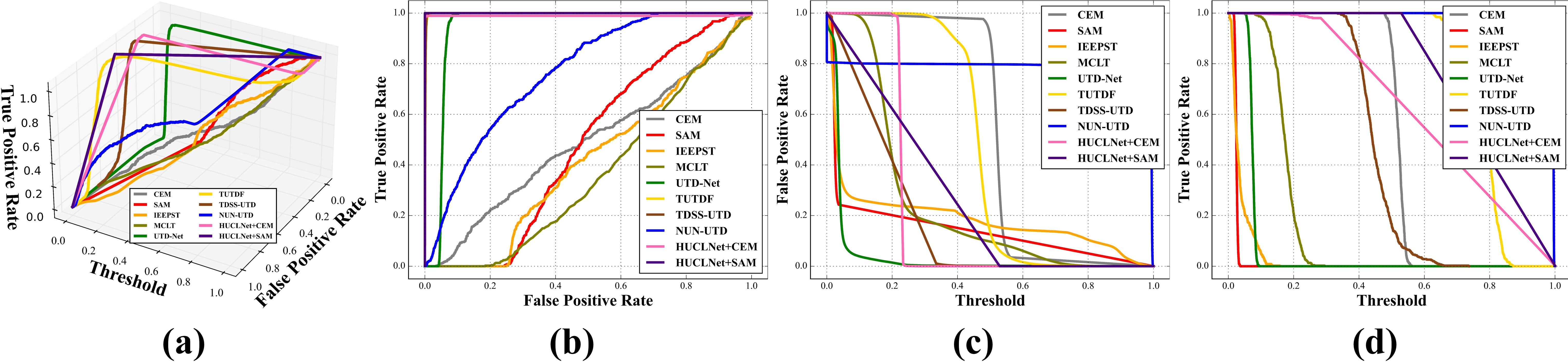}                     
    \caption{ROC curves comparison on ATR2-HUTD Lake Scene2. (a) 3-D ROC curve. (b) 2-D ROC curve of $(P_d, P_f)$. (c) 2-D ROC curve of $(P_f, \tau)$. (d) 2-D ROC curve of $(P_d, \tau)$.}                                  
    \label{fig:C2-2}    
\end{figure*}
Advanced land-cover detection methods, including IEEPST and MCLT, also underperform in underwater environments. IEEPST struggles to suppress background interference, particularly when water column spectral signatures overlap with target signatures in the ATR2-HUTD River sub-dataset. While MCLT leverages contrastive learning for feature enhancement, it shows reduced sensitivity to small or low-reflectance targets, hindered by the nonlinearities and spectral noise typical of underwater HSI data. These results underscore the necessity of specialized techniques for HUTD.

Among SOTA HUTD methods, UTD-Net demonstrates notable improvements by effectively unmixing target-water mixed pixels. However, it faces challenges with background interference in scenes with extensive non-target bottom areas, leading to high false positive rates. NUN-UTD improves target identification by preserving weak target spectral signals, yet remains susceptible to background interference when spectral characteristics of the background resemble those of the target, leading to false positives in spectrally overlapping environments.

Physical-based methods, such as TUTDF and TDSS-UTD, enhance background suppression using underwater imaging models and predicted depth values. However, TUTDF's performance declines in complex environments due to depth estimation inaccuracies, leading to inconsistent detection. Similarly, TDSS-UTD struggles in environments with substantial depth variation, such as the ATR2-HUTD River dataset, where depth errors degrade detection accuracy. Variations in underwater imaging mechanisms between deep and nearshore scenes further limit their effectiveness.

In contrast, HUCLNet-based methods consistently outperform the alternatives. By integrating instance-level and prototype-level contrastive learning, these methods effectively detect faint and deeply submerged targets with minimal false positives, enhancing background suppression and detection accuracy. HUCLNet+CEM and HUCLNet+SAM show resilience to spectral variability, capturing subtle target features while maintaining clear target-background separation, even under significant underwater bottom interference. These methods provide the most comprehensive target coverage and background suppression in challenging environments, such as the ATR2-HUTD River dataset, demonstrating the superior effectiveness of HUCLNet in mitigating spectral variability and improving detection accuracy.
\par 
\textbf{(\romannumeral2) ROC Curves:} Subjective analysis of detection maps may be insufficient for comprehensive evaluation. Therefore, 3-D ROC curves and their 2-D projections: ($P_d$, $P_f$), ($P_d$, $\tau$), and ($P_f$, $\tau$) were used to objectively assess detection performance on the ATR2-HUTD dataset, enabling a detailed evaluation of detection efficiency, target preservation, and background suppression. 
The ROC curves of ATR-HUTD-Lake sub-dataset are provided in Figs~\ref{fig:C2-1} to~\ref{fig:C2-2}, while those of the ATR-HUTD-River and ATR-HUTD-Sea sub-datasets are provided in the supplementary material.
\par
Figs.~\ref{fig:C2-1} (a) to~\ref{fig:C2-2} (a) show the 3-D ROC curves, highlighting the relationship between the true positive rate ($P_d$), false alarm probability ($P_f$), and detection threshold ($\tau$). HUCLNet+CEM and HUCLNet+SAM consistently outperform other methods, exhibiting higher $P_d$ and lower $P_f$ over a wide range of $\tau$, demonstrating superior adaptability.
\par
Figs.~\ref{fig:C2-1} (b) to~\ref{fig:C2-2} (b) present the 2-D ROC curves of ($P_d$, $P_f$). HUCLNet-based methods occupy the top-left region, indicating superior detection accuracy. In contrast, traditional HTD methods, such as CEM and SAM, struggle to balance $P_d$ and $P_f$, particularly for targets with varying spectral properties. Although advanced HTD and SOTA HUTD methods show moderate performance, they fail to suppress false alarms in complex river environments, compromising detection accuracy.

Figs.~\ref{fig:C2-1}(c) to~\ref{fig:C2-2}(c) depict the 2-D ROC curves of ($P_f$, $\tau$), assessing background suppression. NUN-UTD shows high $P_f$ across thresholds, indicating poor background-target discrimination. While methods like MCLT and TUTDF show some improvement, they still struggle with high false alarm rates due to spectral overlap. \textbf{UTD-Net performs well in background suppression but largely by classifying all pixels as background}, as reflected in detection maps (Figs.~\ref{fig:C1-1} to~\ref{fig:C1-2}) and AUC$_{P_{d}, \tau}$ values (Tabs.~\ref{auc_lake} to~\ref{auc_sea}). In comparison, HUCLNet+CEM and HUCLNet+SAM exhibit superior background suppression with low $P_f$ and high AUC$_{P_{d}, \tau}$ values.

Figs.~\ref{fig:C2-1}(d) to~\ref{fig:C2-2}(d) present the 2-D ROC curves of ($P_d$, $\tau$), evaluating target preservation. Traditional methods, such as SAM, show significant drops in $P_d$ as $\tau$ increases, indicating poor target preservation. Advanced HTD and SOTA HUTD methods, such as MCLT and TDSS-UTD, show some improvement but still lag behind NUN-UTD and TUTDF. However, \textbf{the improved performance of NUN-UTD and TUTDF primarily results from misclassifying all pixels as targets}, as shown by high false alarm rates in detection maps (Figs.~\ref{fig:C1-1} to~\ref{fig:C1-2}) and increased AUC$_{P_{f}, \tau}$ values. In contrast, HUCLNet+CEM and HUCLNet+SAM maintain high $P_d$ at lower $\tau$, demonstrating robust and reliable target preservation.
\par
\begin{table*}[!t] 
    \centering
    \footnotesize   
    \caption{Quantitative comparison results on the ATR2-HUTD-Lake Sub-dataset. The best and second best results are in \textbf{bold} and with \underline{underline}.} \label{auc_lake}
    \renewcommand{\arraystretch}{1.5}
    \setlength{\tabcolsep}{1.85mm}
    \scalebox{0.875}
    {
        \begin{tabular}{ccccccccccc}
            \hline
            \multirow{2.4}{*}{\textbf{Method}} & \multicolumn{5}{c}{\cellcolor{tablecolor7!60}\textbf{ATR2-HUTD-Lake Scene1}}       & \multicolumn{5}{c}{\cellcolor{tablecolor8}\textbf{ATR2-HUTD-Lake Scene2}}       \\ \cmidrule(lr){2-6} \cmidrule(lr){7-11}
                                    & $\text{AUC}_{( \mathrm{P}_{\mathrm{d}},\mathrm{P}_{\mathrm{f}})}\textcolor{red}{\uparrow }$ & $\text{AUC}_{( \mathrm{P}_{\mathrm{f}}, \tau)}\textcolor{green}{\downarrow }$ & $\text{AUC}_{( \mathrm{P}_{\mathrm{d}},\tau)}\textcolor{red}{\uparrow }$ & $\mathrm{AUC}_{\mathrm{OA}} \textcolor{red}{\uparrow }$ & $\mathrm{AUC}_{\mathrm{SNPR}}\textcolor{red}{\uparrow }$ & $\text{AUC}_{( \mathrm{P}_{\mathrm{d}},\mathrm{P}_{\mathrm{f}})}\textcolor{red}{\uparrow }$ & $\text{AUC}_{( \mathrm{P}_{\mathrm{f}}, \tau)}\textcolor{green}{\downarrow }$ & $\text{AUC}_{( \mathrm{P}_{\mathrm{d}},\tau)}\textcolor{red}{\uparrow }$ & $\mathrm{AUC}_{\mathrm{OA}} \textcolor{red}{\uparrow }$ & $\mathrm{AUC}_{\mathrm{SNPR}}\textcolor{red}{\uparrow }$ \\ \hline
                                    CEM         & 0.671          & 0.250          & 0.258          & 0.678          & 1.028          & 0.489          & 0.524          & 0.520          & 0.485          & 0.994          \\
                                    SAM         & 0.670          & \underline{0.129}    & 0.151          & 0.692          & 1.170          & 0.480          & \underline{0.143}    & 0.025          & 0.362          & 0.172          \\
                                    IEEPST      & 0.424          & 0.204          & 0.075          & 0.295          & 0.369          & 0.417          & 0.187          & 0.036          & 0.266          & 0.193          \\
                                    MCLT        & 0.401          & 0.422          & 0.377          & 0.357          & 0.894          & 0.365          & 0.243          & 0.173          & 0.296          & 0.715          \\
                                    UTD-Net     & 0.846          & \textbf{0.013} & 0.019          & 0.853          & 1.510          & 0.944          & \textbf{0.041} & 0.073          & 0.976          & 1.773          \\
                                    TUTDF       & \underline{0.990}          & 0.634          & \underline{0.726}    & 1.081          & 1.145          & \underline{0.998}          & 0.461          & \underline{0.768}    & 1.306          & 1.667          \\
                                    TDSS-UTD    & 0.964          & 0.215          & 0.369          & 1.117          & 1.712          & \textbf{0.999} & 0.166          & 0.444          & 1.277          & 2.676          \\
                                    NUN-UTD     & 0.758          & 0.913          & \textbf{0.994} & 0.838          & 1.088          & 0.765          & 0.792          & \textbf{0.995} & 0.968          & 1.257          \\
            \rowcolor{tablecolor13!60}HUCLNet+CEM & 0.958    & 0.302          & 0.642          & \underline{1.298}    & \underline{2.126}    & 0.989    & 0.226          & 0.634          & \underline{1.397}    & \underline{2.805}    \\
            \rowcolor{tablecolor14!60}HUCLNet+SAM & \textbf{0.995} & 0.209          & 0.710          & \textbf{1.501} & \textbf{3.393} & \textbf{0.999} & 0.265          & 0.765          & \textbf{1.501} & \textbf{2.891} \\ \hline
        \end{tabular}}
\end{table*}
\begin{table*}[!t] 
    \centering
    \footnotesize   
    \caption{Quantitative comparison results on the ATR2-HUTD-River Sub-dataset. The best and second best results are in \textbf{bold} and with \underline{underline}.} \label{auc_river}
    \renewcommand{\arraystretch}{1.5}
    \setlength{\tabcolsep}{1.85mm}
    \scalebox{0.875}
    {
        \begin{tabular}{ccccccccccc}
            \hline
            \multirow{2.4}{*}{\textbf{Method}} & \multicolumn{5}{c}{\cellcolor{tablecolor9}\textbf{ATR2-HUTD-River Scene1}}       & \multicolumn{5}{c}{\cellcolor{tablecolor10}\textbf{ATR2-HUTD-River Scene2}}       \\ \cmidrule(lr){2-6} \cmidrule(lr){7-11}
                                    & $\text{AUC}_{( \mathrm{P}_{\mathrm{d}},\mathrm{P}_{\mathrm{f}})}\textcolor{red}{\uparrow }$ & $\text{AUC}_{( \mathrm{P}_{\mathrm{f}}, \tau)}\textcolor{green}{\downarrow }$ & $\text{AUC}_{( \mathrm{P}_{\mathrm{d}},\tau)}\textcolor{red}{\uparrow }$ & $\mathrm{AUC}_{\mathrm{OA}} \textcolor{red}{\uparrow }$ & $\mathrm{AUC}_{\mathrm{SNPR}}\textcolor{red}{\uparrow }$ & $\text{AUC}_{( \mathrm{P}_{\mathrm{d}},\mathrm{P}_{\mathrm{f}})}\textcolor{red}{\uparrow }$ & $\text{AUC}_{( \mathrm{P}_{\mathrm{f}}, \tau)}\textcolor{green}{\downarrow }$ & $\text{AUC}_{( \mathrm{P}_{\mathrm{d}},\tau)}\textcolor{red}{\uparrow }$ & $\mathrm{AUC}_{\mathrm{OA}} \textcolor{red}{\uparrow }$ & $\mathrm{AUC}_{\mathrm{SNPR}}\textcolor{red}{\uparrow }$ \\ \hline
                                    CEM         & 0.746          & 0.280          & 0.300          & 0.765          & 1.070          & 0.650          & 0.544          & 0.553          & 0.659          & 1.016          \\
                                    SAM         & 0.657          & 0.214          & 0.186          & 0.629          & 0.871          & 0.656          & \underline{0.078}    & 0.066          & 0.645          & 0.854          \\
                                    IEEPST      & 0.455          & 0.203          & 0.033          & 0.286          & 0.163          & 0.594          & 0.274          & 0.236          & 0.556          & 0.861          \\
                                    MCLT        & 0.550          & 0.989          & 0.990          & 0.552          & 1.001          & 0.531          & 0.970          & \underline{0.971}    & 0.533          & 1.002          \\
                                    UTD-Net     & \underline{0.843}          & \underline{0.080}    & 0.096          & 0.860          & 1.209          & \underline{0.889}          & \textbf{0.075} & 0.088          & \underline{0.903}          & 1.176          \\
                                    TUTDF       & 0.568          & 0.822          & \underline{0.824}    & 0.570          & 1.003          & 0.659          & 0.356          & 0.363          & 0.667          & 1.022          \\
                                    TDSS-UTD    & 0.402          & 0.438          & 0.415          & 0.379          & 0.948          & 0.539 & 0.179          & 0.174          & 0.534          & 0.974          \\
                                    NUN-UTD     & 0.632          & 0.968          & \textbf{0.999} & 0.663          & 1.032          & 0.503          & 0.977          & \textbf{0.980} & 0.505          & 1.002          \\
            \rowcolor{tablecolor13!60}HUCLNet+CEM & 0.794    & 0.353          & 0.518          & \underline{0.959}    & \underline{1.468}    & 0.753    & 0.354          & 0.481          & 0.880    & \underline{1.360}    \\
            \rowcolor{tablecolor14!60}HUCLNet+SAM & \textbf{0.966} & \textbf{0.055} & 0.175          & \textbf{1.086} & \textbf{3.206} & \textbf{0.924} & 0.178          & 0.327          & \textbf{1.073} & \textbf{1.837} \\ \hline
        \end{tabular}}
\end{table*}
\begin{table*}[!t] 
    \centering
    \footnotesize   
    \caption{Quantitative comparison results on the ATR2-HUTD-Sea Sub-dataset. The best and second best results are in \textbf{bold} and with \underline{underline}.} \label{auc_sea}
    \renewcommand{\arraystretch}{1.5}
    \setlength{\tabcolsep}{1.85mm}
    \scalebox{0.875}
    {
        \begin{tabular}{ccccccccccc}
            \hline
            \multirow{2.4}{*}{\textbf{Method}} & \multicolumn{5}{c}{\cellcolor{tablecolor11}\textbf{ATR2-HUTD-Sea Scene1}}       & \multicolumn{5}{c}{\cellcolor{tablecolor12!50}\textbf{ATR2-HUTD-Sea Scene2}}       \\ \cmidrule(lr){2-6} \cmidrule(lr){7-11}
                                    & $\text{AUC}_{( \mathrm{P}_{\mathrm{d}},\mathrm{P}_{\mathrm{f}})}\textcolor{red}{\uparrow }$ & $\text{AUC}_{( \mathrm{P}_{\mathrm{f}}, \tau)}\textcolor{green}{\downarrow }$ & $\text{AUC}_{( \mathrm{P}_{\mathrm{d}},\tau)}\textcolor{red}{\uparrow }$ & $\mathrm{AUC}_{\mathrm{OA}} \textcolor{red}{\uparrow }$ & $\mathrm{AUC}_{\mathrm{SNPR}}\textcolor{red}{\uparrow }$ & $\text{AUC}_{( \mathrm{P}_{\mathrm{d}},\mathrm{P}_{\mathrm{f}})}\textcolor{red}{\uparrow }$ & $\text{AUC}_{( \mathrm{P}_{\mathrm{f}}, \tau)}\textcolor{green}{\downarrow }$ & $\text{AUC}_{( \mathrm{P}_{\mathrm{d}},\tau)}\textcolor{red}{\uparrow }$ & $\mathrm{AUC}_{\mathrm{OA}} \textcolor{red}{\uparrow }$ & $\mathrm{AUC}_{\mathrm{SNPR}}\textcolor{red}{\uparrow }$ \\ \hline
                                    CEM         & 0.805          & 0.309          & 0.349          & 0.845          & 1.128           & 0.845          & 0.332          & 0.351          & 0.864          & 1.057          \\
                                    SAM         & 0.866          & 0.125    & 0.188          & 0.929          & 1.503           & 0.819          & 0.099          & 0.033          & 0.753          & 0.333          \\
                                    IEEPST      & 0.850          & 0.252          & 0.363          & 0.961          & 1.441           & 0.580          & 0.326          & 0.269          & 0.523          & 0.826          \\
                                    MCLT        & 0.895          & 0.980          & \underline{0.994} & 0.909          & 1.014           & 0.317          & 0.953          & \underline{0.944}    & 0.309          & 0.991          \\
                                    UTD-Net     & 0.762          & \underline{0.050}          & 0.083          & 0.796          & 1.682           & 0.774          & \textbf{0.043} & 0.070          & 0.801          & 1.634          \\
                                    TUTDF       & 0.952          & 0.841          & 0.872          & 0.984          & 1.037           & 0.903          & 0.426          & 0.482          & 0.959          & 1.131          \\
                                    TDSS-UTD    & 0.861          & 0.310          & 0.371          & 0.923          & 1.199           & 0.984          & 0.218          & 0.425          & 1.192          & 1.948          \\
                                    NUN-UTD     & \underline{0.979}    & 0.534          & \textbf{0.999} & \textbf{1.445} & 1.872           & 0.975          & 0.959          & \textbf{0.984} & 0.999          & 1.025          \\
            \rowcolor{tablecolor13!60}HUCLNet+CEM & 0.972          & 0.133          & 0.569          & \underline{1.409}    & \underline{4.284}     & \underline{0.987}    & 0.111          & 0.401          & \underline{1.287}    & \underline{3.620}    \\
            \rowcolor{tablecolor14!60}HUCLNet+SAM & \textbf{0.985} & \textbf{0.019} & 0.325          & 1.292          & \textbf{17.501} & \textbf{0.989} & \underline{0.053}    & 0.474          & \textbf{1.420} & \textbf{8.857} \\ \hline
        \end{tabular}}
\end{table*}
\textbf{(\romannumeral3) AUC Values:} The AUC values for each sub-dataset of the ATR2-HUTD dataset are computed using five key metrics: $\text{AUC}_{( \mathrm{P}_{\mathrm{d}}, \mathrm{P}_{\mathrm{f}})}$, $\text{AUC}_{( \mathrm{P}_{\mathrm{d}}, \tau)}$, $\text{AUC}_{( \mathrm{P}_{\mathrm{f}}, \tau)}$, $\text{AUC}_{SNPR}$, and $\text{AUC}_{OA}$, as detailed in Tabs.~\ref{auc_lake} to~\ref{auc_sea}. These metrics quantitatively assess detection accuracy, target preservation, background suppression, signal-to-noise ratio, and overall performance in varied underwater environments.
\par
\begin{table*}[!ht] 
    \centering
    \footnotesize   
    \caption{Quantitative results of ablation studies on the ATR2-HUTD dataset.} \label{ablation study}
    \renewcommand{\arraystretch}{2}
    \setlength{\tabcolsep}{2.5mm}
    \begin{threeparttable}
        \scalebox{0.975}
        { 
    \begin{tabular}{ccccccc}
        \hline
        \textbf{Module Name}                  & \textbf{Design}                                                      & $\text{AUC}_{( \mathrm{P}_{\mathrm{d}},\mathrm{P}_{\mathrm{f}})}\textcolor{red}{\uparrow }$ & $\text{AUC}_{( \mathrm{P}_{\mathrm{f}}, \tau)}\textcolor{green}{\downarrow }$ & $\text{AUC}_{( \mathrm{P}_{\mathrm{d}},\tau)}\textcolor{red}{\uparrow }$ & $\mathrm{AUC}_{\mathrm{OA}} \textcolor{red}{\uparrow }$ & $\mathrm{AUC}_{\mathrm{SNPR}}\textcolor{red}{\uparrow }$  \\ \hline
        \rowcolor{tablecolor0!50}
        \textbf{HUCLNet}                                          & N/A & 0.943 & 0.188 & 0.502 & 1.258 & 4.446 \\
        \rowcolor{tablecolor1!50}
        \cellcolor{tablecolor1!50}                             & w/o Cluster Refinement Strategy                             & 0.823 & 0.206 & 0.388 & 1.005 & 3.141 \\
        \rowcolor{tablecolor1!50}
        \multirow{-2}{*}{\cellcolor{tablecolor1!50}\textbf{RGC module}}  & w/o Reference Spectrum based Clustering Method & 0.737 & 0.211 & 0.375 & 0.901 & 2.616 \\
        \rowcolor{tablecolor2!50} 
        \cellcolor{tablecolor2!50}                              & w/o Instance-level Contrastive Learning                     & 0.878 & 0.199 & 0.438 & 1.117 & 3.513 \\
        \rowcolor{tablecolor2!50} 
        \cellcolor{tablecolor2!50}                              & w/o Prototype-level Contrastive Learning                    & 0.728 & 0.239 & 0.359 & 0.848 & 2.359 \\
        \rowcolor{tablecolor2!50}
        \cellcolor{tablecolor2!50}                              & w/o Hyperspectral-Oriented Data Augmentation                    & 0.883 & 0.195 & 0.452 & 1.165 & 3.584 \\
        \rowcolor{tablecolor2!50} 
        \multirow{-4}{*}{\cellcolor{tablecolor2!50}\textbf{HLCL module}} & w/o HLCL module$^{1}$                                             & 0.696 & 0.252 & 0.248 & 0.692 & 0.933 \\
        \rowcolor{tablecolor3!50} 
        \textbf{SPL Paradigm}                                          & w/o SPL Paradigm                                            & 0.743 & 0.217 & 0.388 & 0.914 & 2.864 \\ \hline
        \end{tabular}}
        \begin{tablenotes}
            \scriptsize
            \item[1] This experimental design is analogous to the baseline HTD methods, as the RGC module and SPL paradigm are dependent on the HLCL module for functionality.
        \end{tablenotes}
        \end{threeparttable}
\end{table*}
The $\text{AUC}_{( \mathrm{P}_{\mathrm{d}}, \mathrm{P}_{\mathrm{f}})}$ metric, which quantifies the trade-off between the true positive rate ($P_d$) and false alarm probability ($P_f$), is critical for evaluating detection performance. HUCLNet+SAM leads with an average score of 0.976, followed by HUCLNet+CEM at 0.909. Traditional methods, such as SAM (0.701) and MCLT (0.691), underperform significantly, while SOTA HUTD methods like TUTDF and NUN-UTD fall short of HUCLNet-based methods in detection capability.
\par
For background suppression, assessed by $\text{AUC}_{( \mathrm{P}_{\mathrm{f}}, \tau)}$, HUCLNet+SAM achieves the highest performance in the ATR2-HUTD-River Scene1 and ATR2-HUTD-Sea sub-datasets, the most complex nearshore environments. It also demonstrates robust performance across other sub-datasets. In contrast, SOTA HUTD methods, including TUTDF and NUN-UTD, show elevated values, suggesting overfitting due to high false positive rates.
\par
The $\text{AUC}_{( \mathrm{P}_{\mathrm{d}}, \tau)}$ metric, assessing target preservation, reveals HUCLNet-based methods performing well, though NUN-UTD leads. This may be attributed to the HLCL module in HUCLNet, which compromises target-background feature separation, impacting target preservation. Additionally, NUN-UTD's higher false positive rate boosts $P_d$ but hinders background suppression.
\par
The $\text{AUC}_{OA}$ metric, combining $\text{AUC}_{( \mathrm{P}_{\mathrm{d}}, \mathrm{P}_{\mathrm{f}})}$, $\text{AUC}_{( \mathrm{P}_{\mathrm{d}}, \tau)}$, and $\text{AUC}_{( \mathrm{P}_{\mathrm{f}}, \tau)}$, further emphasizes HUCLNet's superiority. HUCLNet+SAM achieves the highest average score of 1.312, with HUCLNet+CEM following at 1.205. In contrast, traditional and SOTA HUTD methods score between 0.492 and 0.928, underscoring HUCLNet's effectiveness in background suppression, target preservation, and detection accuracy in complex nearshore environments.
\par
Finally, the $\text{AUC}_{SNPR}$ metric, which measures robustness under varying signal-to-noise ratios, underscores HUCLNet+SAM's superior performance, achieving the highest scores across all sub-datasets, including 17.501 in ATR2-HUTD-Sea Scene1. HUCLNet+CEM consistently ranks second, while traditional HTD and SOTA HUTD methods show lower scores, indicating reduced robustness in fluctuating signal conditions.
\par
\subsection{Ablation Studies}\label{sec:4.4}
To evaluate the efficacy of each component in our method, we conducted ablation studies on the ATR2-HUTD dataset. These studies aim to confirm that the observed improvements stem not only from the increased number of parameters but also from the architectural design, which enhances the HUTD task. The HUCLNet framework is divided into three components for experimental validation. 
Corresponding results are presented in Tab.~\ref{ablation study}.
\par
\textbf{(\romannumeral1) Analysis of the RGC Module:} We validate the RGC with the following designs: 
\begin{itemize}
    \item \textbf{w/o Cluster Refinement Strategy:} This design excludes the cluster refinement strategy, relying solely on the reference spectrum-based clustering method. 
    \item \textbf{w/o Reference Spectrum-based Clustering:} This design omits the reference spectrum-based clustering approach from the RGC module.
\end{itemize}
\par
Without the cluster refinement strategy, the RGC module directly uses the original clustering results, often misclassifying pixels and negatively impacting prototype-level learning. As seen in Tab.~\ref{ablation study}, this leads to lower average metric values compared to the full HUCLNet-based methods, demonstrating the importance of refined pseudo-labels. Removing the RGC module entirely, the HLCL module uses pixel instances from the original HSI for instance-level contrastive learning, focusing on individual pixel spectra and neglecting the target-background relationships. Performance improves slightly over baseline HTD methods but remains significantly inferior to complete HUCLNet-based methods, highlighting the critical role of the RGC module in providing reliable pseudo-labels.
\par
\begin{figure*}[!t]                 
    \centering                    
    \includegraphics[width=2\columnwidth]{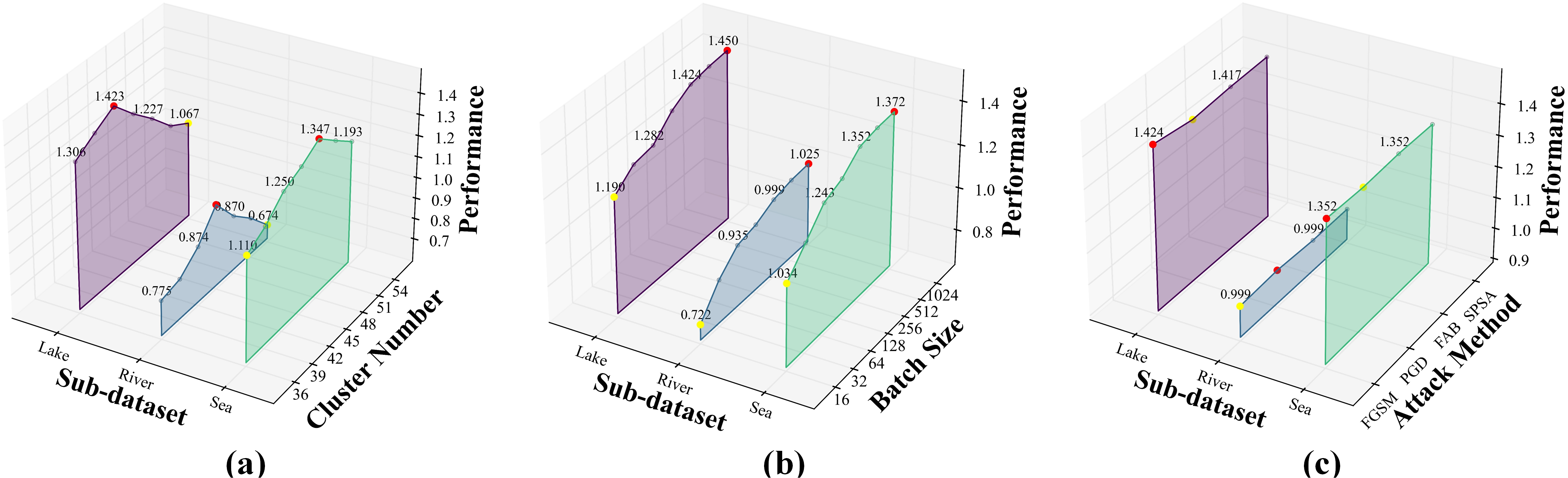}                     
    \caption{Parameter analysis results on the ATR2-HUTD dataset. (a) Number of clusters in the RCG module; (b) Batch size in the HLCL module; (c) Attack method in the HLCL module. Red and yellow points indicate the maximum and minimum values, respectively.}                 
    \label{fig:C4-1}    
\end{figure*}
\textbf{(\romannumeral2) Analysis of the HLCL Module:} We evaluate the HLCL module with the following designs:
\begin{itemize}
    \item \textbf{w/o Instance-level Contrastive Learning:} This design removes instance-level contrastive learning, relying only on refined cluster labels from the RGC module.
    \item \textbf{w/o Prototype-level Contrastive Learning:} This design removes prototype-level contrastive learning, retaining only instance-level contrastive learning.
    \item \textbf{w/o Hyperspectral-Oriented Data Augmentation:} This design removes hyperspectral-specific data augmentation from the HLCL module.
    \item \textbf{w/o HLCL Module:} This design excludes the entire HLCL module.
\end{itemize}
\par
According to Tab.~\ref{ablation study}, we can draw the following conclusions.
When the HLCL module operates without instance-level contrastive learning, HUCLNet relies solely on the cluster labels, leading to performance degradation. However, prototype-level contrastive learning alone still outperforms baseline HTD methods, emphasizing the importance of target-background separability. The removal of prototype-level contrastive learning results in poorer performance compared to the instance-level design, indicating its greater impact on separability. When hyperspectral-oriented data augmentation is excluded, traditional augmentation methods lead to observable performance degradation, confirming the importance of hyperspectral-specific augmentation in enhancing feature discriminability and HUCLNet's performance. Finally, removing the HLCL module entirely reduces HUCLNet to baseline HTD methods, resulting in substantial performance loss, reinforcing the HLCL module's primary contribution to performance improvement.
\par
\textbf{(\romannumeral3) Analysis of the SPL Paradigm:} We evaluate the SPL paradigm with the following design: 
\begin{itemize}
    \item \textbf{w/o SPL Paradigm:} This design trains the model using the traditional self-supervised learning framework, which consists of a single reliable-guided clustering step followed by hybrid-level contrastive learning.
\end{itemize}
\par
Without the SPL paradigm, inaccurate clustering due to limited spectral discriminability hinders contrastive learning effectiveness, resulting in error propagation and performance degradation. Tab.~\ref{ablation study} confirms that the SPL paradigm significantly enhances HUCLNet's performance, underscoring the importance of the self-paced strategy in guiding model training and improving detection accuracy.
\par
\subsection{Parameter Analysis}\label{sec:4.5}
The key hyperparameters of the HUCLNet architecture, including the number of clusters in the RGC module, batch size, and attack method in the HLCL module, were analyzed through experiments on the ATR2-HUTD dataset. The results, primarily focusing on the $\text{AUC}_{\text{OA}}$ metric, are presented in Fig.~\ref{fig:C4-1}, as it is the most critical indicator of overall detection performance.
\par
\textbf{(\romannumeral1) Number of Clusters in the RGC Module:} The number of clusters in the RGC module plays a crucial role in clustering accuracy and overall HUCLNet performance. The number of clusters was varied between 30 and 48, with a step size of 3 (Fig.~\ref{fig:C4-1} (a)). Performance improves with an increasing number of clusters up to an optimal point, after which it deteriorates due to over-segmentation, where target pixels are fragmented into multiple clusters. This fragmentation hinders prototype-level contrastive learning, leading to inconsistent target representations. For the ATR2-HUTD Lake, River, and Sea sub-datasets, the optimal number of clusters was 36, 39, and 42, respectively. Even with suboptimal cluster numbers, HUCLNet outperforms baseline methods.

\textbf{(\romannumeral2) Batch Size in the HLCL Module:} The batch size in the HLCL module is another critical parameter affecting HUCLNet performance. Varying the batch size from 32 to 512 with a step size of 64, results (Fig.~\ref{fig:C4-1} (b)) show that larger batch sizes generally improve performance by increasing the number of negative samples, enhancing feature discriminability. This is consistent with prior work~\cite{Chen2020}, which indicates that larger batch sizes benefit contrastive learning. However, performance gains plateau at higher batch sizes, and larger sizes impose greater memory and computational demands. A batch size of 256 provides an optimal balance between performance and resource usage across all ATR2-HUTD sub-datasets.

\textbf{(\romannumeral3) Attack Method in the HLCL Module:} The choice of attack method in the HLCL module influences the generation of adversarial samples for contrastive learning. Four attack methods—FGSM~\cite{GoodfellowSS14}, PGD~\cite{MadryMSTV18}, FAB~\cite{Croce020}, and SPSA~\cite{SPSA}—were tested with a perturbation limit of $\epsilon = 0.1$. As shown in Fig.~\ref{fig:C4-1} (c), performance across attack methods is similar, suggesting that the specific choice of attack method has minimal impact, as long as the generated adversarial samples are effective. Given its computational efficiency and comparable performance, we adopt the FGSM attack method for HUCLNet.

\section{Conclusion} \label{sec:6}
This paper addresses the challenge of detecting underwater targets in nearshore environments, where severe water attenuation distorts spectral characteristics. We propose a UAV-borne hyperspectral target localization strategy, supported by the ATR2-HUTD benchmark dataset, specifically designed for accurate underwater detection. The dataset includes three UAV-borne hyperspectral sub-datasets, each representing distinct underwater scenarios.
To improve detection, we introduce HUCLNet, a hybrid-level contrastive learning framework that integrates reliability-guided clustering and a self-paced learning paradigm, optimized for UAV-borne hyperspectral imagery.
Extensive experiments on the ATR2-HUTD dataset demonstrate HUCLNet's superior performance across multiple evaluation metrics, including detection accuracy, target preservation, background suppression, signal-to-noise ratio, and overall detection effectiveness, outperforming both traditional and state-of-the-art methods.
Ablation studies and hyperparameter analyses confirm the contributions of each HUCLNet component, providing insights into optimal configurations for maximal performance. Future work will explore HUCLNet's application in more complex underwater environments and assess its generalization across diverse hyperspectral sensors.

\bibliographystyle{IEEEtran}
\bibliography{IEEEabrv,ref}

\end{document}